\documentclass[journal]{IEEEtran}
\usepackage{graphicx}
\usepackage{bm}
\usepackage{caption}
\usepackage{amsmath}
\usepackage{amssymb}
\usepackage{mathtools}
\usepackage{multirow}
\usepackage{xcolor}
\usepackage{subcaption}
\usepackage{yfonts}
\usepackage[ruled]{algorithm2e}
\usepackage[margin=0.75in,bottom=0.8in]{geometry}
\usepackage{titling}
\date{}
\newtheorem{lemma}{Lemma}
\newtheorem{assumption}{Assumption}
\newtheorem{remark}{Remark}
\newtheorem{prob}{Problem}
\newtheorem{property}{Property}

\title{Path Planning for Continuum Rods Using Bernstein Surfaces}
\author{Maxwell Hammond, Venanzio Cichella, Amirreza F. Golestaneh, Caterina Lamuta}

\begin{document}
\maketitle
\begin{abstract}
This paper presents a method for optimal motion planning of continuum robots by employing Bernstein surfaces to approximate the system's dynamics and impose complex constraints, including collision avoidance. The main contribution is the approximation of infinite-dimensional continuous problems into their discrete counterparts, facilitating their solution using standard optimization solvers. This discretization leverages the unique properties of Bernstein surface, providing a framework that extends previous works which focused on ODEs approximated by Bernstein polynomials. Numerical validations are conducted through several numerical scenarios. The presented methodology offers a promising direction for solving complex optimal control problems in the realm of soft robotics. 
\end{abstract}

\section{Introduction}
Research in the field of robotics continues to introduce new materials and methods for creating complex systems capable of unique actuation with large numbers of degrees of freedom (DOF) \cite{greco2022evolution,hammond2023bioinspired}. The effects of these advancements are of particular note for the construction of continuum robotic systems. This class of robots offers many potential advantages over discrete counterparts owing to near infinite DOF, particularly in applications which require navigation through dense environments, e.g. in invasive surgeries or underwater exploration  \cite{singh2014continuum}. Even so, several obstacles remain to ensure safe implementation of these devices. While some challenges may be addressed by research in smart materials and design, other efforts must be made to close existing gaps in modeling, path planning, and control \cite{da2020challenges,burgner2015continuum}. 

Among the aforementioned challenges, modeling is a core issue for which a consensus on a solution has not been reached \cite{santina2023model}.  Employing the strongest assumptions among existing approaches are models which discretize a continuum system to a chain of rigid links to simplify the dynamics \cite{chirikjian1994modal,venkiteswaran2019shape,khoshnam2013pseudo}. While this method may work for simple cases with good feedback, it is not ideal for path planning or less trivial applications. Other approaches model the central cord of the continuum system as a one-dimensional rod element parameterized by its arc length. Piecewise constant curvature (PCC) models are one such example which have seen wide use \cite{zhong2021bending,della2020improved,falkenhahn2015model}, discretizing the system and assuming constant curvature along the arc of each rod segment. This assumption eliminates the need for a continuous spatial derivative in modeling, and allows the dynamics to be expressed as a system of time dependant ordinary differential equations (ODEs); however, the simplification eliminates the model's ability to capture deformation which a robot may be capable of, again limiting uses. The Cosserat rod model is another example of a one-dimensional modeling approach which has no specific assumptions, allowing it to fully describe a system's pose and all modes of deformation \cite{renda20123d,janabi2021cosserat,till2019real}. This model takes the form of a set of partial differential equations (PDEs), leading to a number of challenges including a need for numerical solution methods and complex controller derivation. Nevertheless, this model has been used in the formulation of several controllers, including energy shaping control \cite{chang2023energy,chang2021controlling}, infinite dimensional state feedback control \cite{zheng2022pde}, sliding mode control \cite{alqumsan2019robust}, and others \cite{wang2022sensory,till2017elastic}. On the other hand, path planning with Cosserat is not well explored and only a few examples can be found in literature. Hermansson et al. \cite{hermansson2021quasi} employ a discretized Cosserat model for quasi-static path optimization. In contrast, the works in \cite{hermansson2013automatic} and \cite{hermansson2016automatic} leverage Cosserat dynamics to plan the routing of flexible components within manufacturing workflows. Additionally, Messer et al. \cite{messer2022ctr} adopt a simplified kinematic representation of Cosserat for path planning of concentric tube robots (CTR). In these cases, sacrifices are made in either computational efficiency or accuracy, leaving room for significant improvement.

Recently, the use of Bernstein polynomials (or Bezier curves) has been proposed to efficiently approximate dynamics in optimal control problems (OCPs), allowing them to be solved as nonlinear programming problems (NLPs) \cite{kielas2022bernstein,cichella2017optimal}. The Bernstein basis is known to be numerically stable, and has a number of geometric properties which can be exploited for constraint enforcement and obstacle avoidance \cite{kielas2019bebot}. Additionally, the computational efficiency of this method has proven sufficient for enabling real time path planning in multi-agent missions \cite{cichella2020optimal}. Univariate Bernstein polynomials are well suited for approximating ODEs; however, the Bernstein basis can be extended to surfaces for approximating PDEs over additional dimensions. While the use of Bernstein surfaces in path planning applications is novel to the best of the author's knowledge, the use of other polynomial bases in PDE based OCPs has been explored to some extent. Power series \cite{meurer2011finite}, proper orthogonal decomposition \cite{radmanesh2021pde}, and fifth degree polynomial curves \cite{liu2021path} are examples of such polynomial bases which have been used for solutions in cases of multi-agent path planning, or planning in complex flow environments. Even so, the geometric properties and efficiency of Bernstein surfaces make them ideal in the case of Cosserat rod path planning, as they can guarantee safe trajectories and constraint satisfaction at low orders of approximation. This is critical in soft robotic applications, which may be using delicate actuators or require quick responses to dynamic environments.

In this paper, a method for Cosserat rod path planning is described using Bernstein surfaces to approximate the system's kinematic constraints in the formulation of an OCP. The paper is organized as follows: Section \ref{sec:cos} describes the Cosserat rod model and culminates with the formulation of the optimal motion planning problem for this model. Section \ref{sec:bern} introduces the Bernstein basis and reformulates the optimal motion planning problem to an NLP problem. Section \ref{sec:res} discusses some numerical results. Section \ref{sec:conc} provides conclusion.

\section{Problem Formulation} \label{sec:cos}
\begin{figure}
    \centering
    \includegraphics[width=0.275\textwidth]{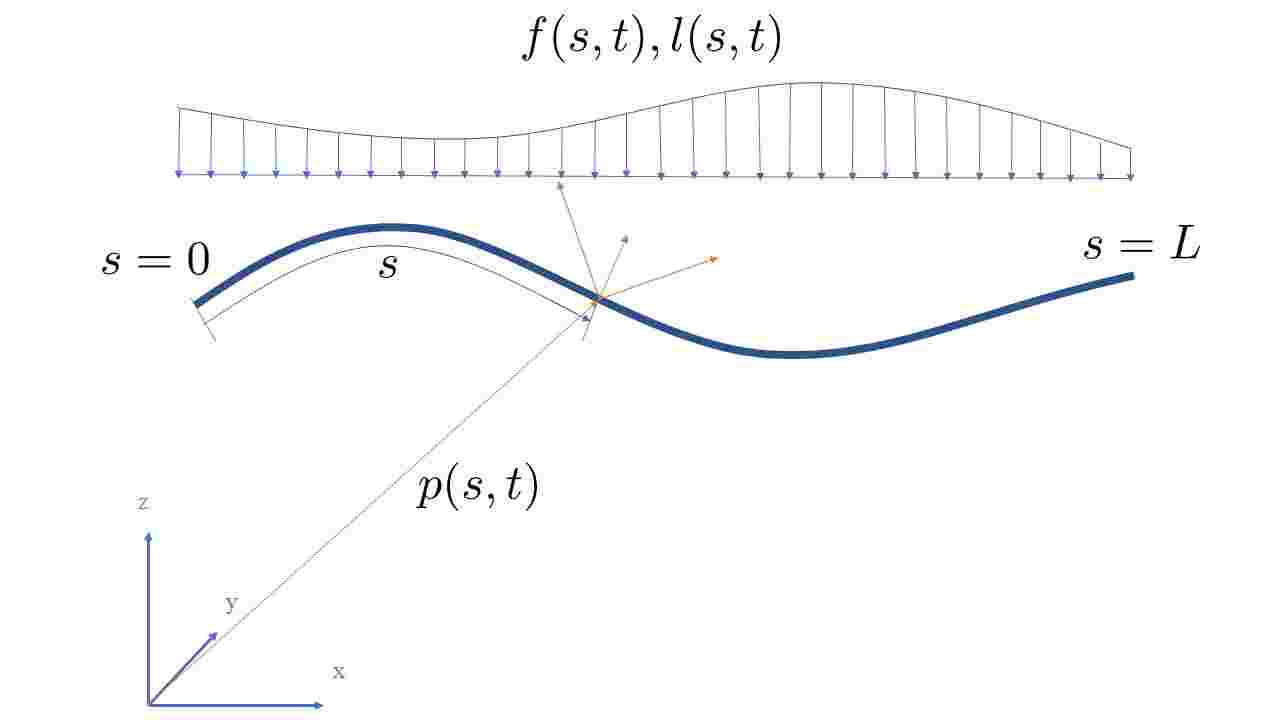}
    \caption{A Cosserat rod of length $L$ being acted on by external distributed forces, $f(s,t)$, and moments, $l(s,t)$ shown in a global frame.}
    \label{fig:cosserat}
\end{figure}

The Cosserat rod model captures the dynamics of a slender system as a finite dimensional continuum rod. The dynamics of the position and orientation, i.e., the pose, of the rod along its arc length, $s\in[0,L]$, and for all time, $t\in[0,T]$, is governed by the following partial differential equations: 
\begin{table}
\caption{Cosserat Rod Variables}
\label{table:var}
\centering
\begin{tabular}{|c||c|c|c|} 
    \hline
    \multicolumn{4}{|c|}{Cosserat Rod Variable List} \\
    \hline
    Variable & Unit & Frame & Description \\
    \hline
    $s$ & $m$ & (-) & arc length \\
    $t$ & $s$ & (-) & time \\
    $\textbf{p}(s,t)$ & $m$ & global & position \\
    $\textbf{R}(s,t)$ & (-) & global & orientation \\
    $\textbf{q}(s,t)$ & $m/s$ & local & linear velocity \\
    $\boldsymbol{\omega}(s,t)$ & $s^{-1}$ & local & angular velocity \\
    $\textbf{v}(s,t)$ & (-) & local & linear stretch \\ 
    $\textbf{u}(s,t)$ & $m^{-1}$ & local & angular strain \\
    $\textbf{n}(s,t)$ & $N$ & global & internal forces \\
    $\textbf{m}(s,t)$ & $Nm$ & global & internal moments \\
    $\textbf{f}(s,t)$ & $N/m$ & global & external distributed forces \\
    $\textbf{l}(s,t)$ & $Nm/m$ & global & external distributed moments \\
    $\rho$ & $Kg/m^3$ & (-) & material density \\
    $A$ & $m^2$ & (-) & material cross sectional area \\
    $\textbf{J}$ & $m^4$ & (-) & mass moment of inertia \\
    \hline
\end{tabular}
\end{table}
\begin{equation} \label{eq:cossdyn}
\begin{aligned}
    \textbf{p}_s &= \textbf{Rv},\quad 
    \textbf{p}_t = \textbf{Rq},\\
    \textbf{R}_s &= \textbf{R}\hat{\textbf{u}},\quad
    \textbf{R}_t = \textbf{R}\hat{\boldsymbol{\omega}},\\
    \textbf{n}_s &= \rho A\textbf{R}(\boldsymbol{\omega}\textbf{q}+\textbf{q}_t)-\textbf{f},\\
    \textbf{m}_s &= \rho\textbf{R}(\hat{\boldsymbol{\omega}}\textbf{J}\boldsymbol{\omega}+\textbf{J}\boldsymbol{\omega}_t)-\hat{\textbf{p}}_s\textbf{n}-\textbf{l},\\
    \textbf{q}_s &= \textbf{v}_t-\hat{\textbf{u}}\textbf{q}+\hat{\boldsymbol{\omega}}\textbf{v},\\
    \boldsymbol{\omega}_s &= \textbf{u}_t-\hat{\textbf{u}}\boldsymbol{\omega}.
\end{aligned}
\end{equation}
The subscripts $x_s$ and $x_t$ denote the partial derivatives with respect to $s$ and $t$, respectively. The \textit{hat} operator provides the mapping between $\mathbb{R}^3$ to $so(3)$. A description of the variables associated with this model can be found in Table \ref{table:var} and a visual representation can be seen in Figure \ref{fig:cosserat}. To fully describe a physical system with this model, a constitutive law must also be chosen to describe the relationship between strains and internal forces. This law is material dependant; however it is often chosen as a simple linear model 
\begin{equation}
\begin{aligned}
    \textbf{n} &= \textbf{R}\left[\textbf{K}_{se}(\textbf{v}-\textbf{v}_0)+\textbf{B}_{se}\textbf{v}_t\right], \\
    \textbf{m} &= \textbf{R}\left[\textbf{K}_{bt}(\textbf{u}-\textbf{u}_0)+\textbf{B}_{bt}\textbf{u}_t\right], \\
\end{aligned}
\end{equation} 
where the matrices $\textbf{K}_{se}$ and $\textbf{B}_{se}$ relate internal forces to transnational strain and its time rate, and  $\textbf{K}_{bt}$ and $\textbf{B}_{bt}$ perform similar function for the internal moment's relation to angular strain.

Next, we assume the existence of a tracking algorithm that stabilizes the dynamics of the rod.
\begin{assumption} \label{asm:tracking}
Consider a continuum rod with motion governed by Equation \eqref{eq:cossdyn}. Let $\mathbf{p}_{\text{ref}}(s,t)$ and $\mathbf{R}_{\text{ref}}(s,t)$ denote the reference pose to be tracked by the rod, which is assumed to have bounded linear and angular strains, as well as bounded linear and angular velocities.
We assume that the rod has a tracking controller for \(\mathbf{f}\) and \(\mathbf{l}\) such that the generalized error vector \(\mathbf{e} = [\mathbf{e}_p^\top \, e_R]^\top\), with \(\mathbf{e_p} = \mathbf{p} - \mathbf{p}_\text{ref}\) and \(e_R = \text{trace} (\mathbf{I}-\mathbf{R}^\top\mathbf{R}_{\text{ref}})\), is uniformly bounded. Specifically, for some positive constant \(c\), and any \(a \in (0,c)\), there exists \(\rho = \rho(a) > 0\) where
\[ ||\mathbf{e}(s,0)|| \leq a \implies ||\mathbf{e}(s,t)|| \leq \rho , \quad \forall t\geq 0 . \]
\end{assumption}
In the equations mentioned above, the norm \(||\cdot||\) is assumed to be the Euclidean norm. 


\begin{remark}
Assumption \ref{asm:tracking} is frequently adopted in the path planning literature, see for example, \cite{messer2022ctr,gasparetto2015path}. 
In fact, multi-loop architectures, which utilize an inner-loop controller to stabilize the robot's dynamics and outer-loop algorithms for high-level path planning, are often preferred. This design offloads some computational demands from the motion planning, facilitating the prioritization of higher-level objectives. This consequently allows for rapid computations and near real-time responses. 
\end{remark}

\begin{remark}
Literature provides several examples of inner loop controllers that meet the criteria set by Assumption \ref{asm:tracking}, e.g., \cite{chang2023energy,till2017elastic}.
\end{remark}

Under Assumption \ref{asm:tracking}, our investigation narrows to motion planning, with an emphasis on generating the reference pose \((\mathbf{p}_{\text{ref}}(s,t),\mathbf{R}_{\text{ref}}(s,t))\). For simplicity in notation, we will refer to the reference position and orientation as \(\mathbf{p}(s,t)\) and \(\mathbf{R}(s,t)\), omitting the subscript ``\text{ref}''. Occasionally, we may omit the explicit dependency on $s$ and $t$ and simply refer to them as $\mathbf{p}$ and $\mathbf{R}$. Next, we detail a mathematical framework for the goals the motion planner should achieve.


\textbf{Feasibility constraints.} Primarily, the motion planner must consider dynamics and {feasibility constraints} to ensure that the robot can track the generated pose. Motivated by Assumption \ref{asm:tracking}, we say that a pose \((\mathbf{p},\mathbf{R})\) governed by  
\begin{equation}
    \begin{aligned}
        &\mathbf{p}_s = \mathbf{R}\mathbf{v}, & 
        &\mathbf{p}_t = \mathbf{R}\mathbf{q}, \\
        &\mathbf{R}_s = \mathbf{R}\hat{\mathbf{u}}, &
        &\mathbf{R}_t = \mathbf{R}\hat{\boldsymbol{\omega}},
    \end{aligned}
\label{eq:cosserat2}
\end{equation}
is feasible if its velocities, both spatial and temporal, are bounded:
\begin{equation} \label{eq:vconstraints}
\begin{aligned}
v_{\min}^2 \leq ||\mathbf{v}||^2 &\leq v_{\max}^2, & ||\mathbf{q}||^2 &\leq q_{\max}^2, \\
||{u}||^2&\leq u_{\max}^2, & ||{\omega}||^2 &\leq \omega_{\max}^2, \\
||\mathbf{v}_s||^2 &\leq v_{s,\max}^2, & ||\mathbf{q}_t||^2 &\leq q_{t,\max}^2,
\end{aligned}
\end{equation}
for given \(v_{\max},q_{\max},u_{\max},\omega_{\max} \in \mathbb{R}^+\). 


\textbf{Boundary conditions.} The motion planning algorithm must integrate the {boundary constraints}. Specifically:
the resulting pose should be consistent with the robot's initial states; specific constraints can be set for the final states; the robot may be tethered to a fixed point.
These conditions can be formulated as: 
\begin{equation} \label{eq:boundary} 
\begin{aligned} 
\mathbf{g}_i(\mathbf{p}(s,0),\mathbf{R}(s,0),\mathbf{q}(s,0),\boldsymbol{\omega}(s,0)) & = 0 \\
\mathbf{g}_f(\mathbf{p}(s,T),\mathbf{R}(s,T),\mathbf{q}(s,T),\boldsymbol{\omega}(s,T)) & = 0 \\
\mathbf{g}_0(\mathbf{p}(0,t),\mathbf{R}(0,t),\mathbf{q}(0,t),\boldsymbol{\omega}(0,t)) & = 0 
\end{aligned}
\end{equation} 


\textbf{Obstacle avoidance.} Finally, in the motion planning framework, collision avoiding against obstacles is paramount. To facilitate this, let \( d(\mathbf{p}(s,t), S) \)  be the minimum separation between point \(\mathbf{p}(s,t)\) on the rod and solid \( S \). I.e.,
\begin{equation*}
    d(\mathbf{p}(s,t), S) = \min_{\mathbf{x} \in S} ||\mathbf{p}(s,t) - \mathbf{x}||
\end{equation*}
Here, \( \mathbf{x} \) represents any point within, or on the boundary of, the solid \( S \). 
Building on this, the continuum rod's obstacle avoidance constraint, which ensures it does not penetrate the solid, can be formulated as:
\begin{equation} \label{eq:collisionavoidance}
    d(\mathbf{p}(s,t), S) \geq d_{\text{safe}} > 0 , \qquad s \in [0,L], \, t \in [0,tf]\geq 0
\end{equation}
In this context, \( d_{\text{safe}} \) serves as the requisite safety margin.

\vspace{5px}

With the above setup, the motion planning problem can be formulated as an optimal control problem over PDEs. In particular, the motion planning problem is formally stated as follows. 

\vspace{5px}

\begin{prob} \label{prob:continuous}
\begin{equation}
\min_{\mathbf{p},\mathbf{R},T} \int_{0}^{T} \int_0^L  \ell (\mathbf{p}(s,t),\mathbf{R}(s,t)))\, ds \, dt 
\end{equation}
subject to Equations \eqref{eq:cosserat2}, \eqref{eq:vconstraints}, \eqref{eq:boundary}, and \eqref{eq:collisionavoidance}.
\end{prob}

\vspace{5px}

The running cost \( \ell: \mathbb{R}^3 \times \mathbb{R}^{3 \times 3} \times \mathbb{R} \to \mathbb{R}\), as presented in Problem \ref{prob:continuous}, is contingent upon the specific goals and requirements of the planning algorithm tailored to the particular application. The following is assumed about this function.
\begin{assumption}
\(\ell\) is Lipschitz continuous with respect to its arguments. 
\end{assumption}

Given the complexity of Problem \ref{prob:continuous}, numerical methods need to be deployed to find optimal position, $\mathbf{p}^* : [0,L] \times [0,T]\to\mathbb{R}^3$, and orientation, $\mathbf{R}^* : [0,L] \times [0,T]\to\mathbb{R}^{3 \times 3}$, that solves it. 
In the following section, we present an approximation method that transforms Problem \ref{prob:continuous} into a non-linear programming problem utilizing Bernstein surfaces. This converted problem can subsequently be addressed using readily available optimization software tools.

\section{Bernstein approximation of Problem \ref{prob:continuous}}
\label{sec:bern}
We let the position of the rod be parameterized by an \(m \times n\) Bernstein surface (constructed using Bernstein polynomials in two variables) given by the following tensor product: 
\begin{equation} \label{eq:BSposition}
    \quad \textbf{p}(s,t) = \sum_{i=0}^{m}\sum_{j=0}^{n}\bar{\textbf{p}}_{i,j}^{m,n}B_{i}^{m}(s)B_{j}^{n}(t).
\end{equation}
where $\bar{\textbf{p}}_{i,j}^{m,n}\in \mathbb{R}^{3}$, \(i=0,\ldots,m\), \(j=0,\ldots,n\) are control points and $B_{i}^{m}(s)$ is the Bernstein polynomial basis over the interval $[0,L]$, given by
\begin{equation*} 
    B_{i}^{m}(s)={m \choose i}\frac{s^i\left(L-s\right)^{m-i}}{L^m}, \quad 0 \le i \le m, 
\end{equation*}
where ${m \choose i} = \frac{m!}{i!\left(m-i\right)!}.$ The orders \(m\) and \(n\) denote the degree of the Bernstein surface in the respective \(s\) and \(t\) variables.
The orientation \(\mathbf{R}\) is described using the \(XYZ\) Euler angles \((\phi(s,t),\theta(s,t),\psi(s,t))\). Specifically, the Bernstein surface representing the Euler rotation about the \(X\)-axis is given by
\begin{equation} \label{eq:phi}
    {\phi}(s,t) = \sum_{i=0}^{m}\sum_{j=0}^{n}\bar{\phi}_{i,j}^{m,n}B_{i}^{m}(s)B_{j}^{n}(t),
\end{equation}
while rotations about the \(Y\) and \(Z\) axes, namely \(\theta(s,t)\) and \(\psi(s,t)\), respectively, are analogously represented, with control points denoted as \(\bar{\theta}^{m,n}_{i,j}\) and \(\bar{\psi}^{m,n}_{i,j}\) with \(i=0,\ldots,m\) and \(j = 0,\ldots,n\). 

\begin{remark}
    The use of Euler angles is motivated by their simplicity and the requirement of only three elements to depict the orientation. However, to circumvent singularities that may arise from Euler angle representations, alternative representations like quaternions can be employed without altering our approach.
\end{remark}

With this parameterization, the feasibility constraints introduced in Equations \eqref{eq:cosserat2} and \eqref{eq:vconstraints} can be rewritten as follows:
\begin{equation} \label{eq:vconstraintsder}
\begin{aligned}
v_{\min}^2 \leq ||\mathbf{p}_s||^2 &\leq v_{\max}^2, & ||\mathbf{p}_t||^2 &\leq q_{\max}^2, \\
||\, [\phi_s ,  \theta_s, \psi_s] \, ||^2&\leq u_{\max}^2, & || \, [\phi_t ,  \theta_t, \psi_t] \, ||^2 &\leq \omega_{\max}^2, \\
||\mathbf{p}_{ss}||^2 &\leq v_{s,\max}^2, & ||\mathbf{p}_{tt}||^2 &\leq q_{t,\max}^2,
\end{aligned}
\end{equation}

We now introduce some properties of Bernstein surfaces that aid the computation of the above constraints. 

\begin{property}[Arithmetic operations]
Addition and subtraction between two Bernstein surfaces can be performed directly through the addition and subtraction of their control points. The control points of the Bernstein surface \(y(\cdot,\cdot)\) resulting from multiplication between two Bernstein surfaces, \(g(\cdot,\cdot)\) and \(h(\cdot,\cdot)\) with control points $\bar{g}_{i,j}^{m,n}$ and $\bar{h}_{k,l}^{a,b}$ can be obtained by 
\begin{equation}
    \begin{split}
        \bar{y}_{e,f}^{m+a,n+b} &= \\ \sum_{q=max(0,e-a)}^{min(m,e)}&\sum_{r=max(0,f-b)}^{min(n,f)}  
        \frac{{m \choose q}{n \choose r}{a \choose e-q}{b \choose f-r}}{{m+a \choose e}{n+b \choose f}}\bar{g}_{q,r}^{m,n}\bar{h}_{e-q,f-r}^{a,b}.
    \end{split}
\end{equation}
Note that multiplication will result in a Bernstein surface whose order is the sum of the orders of the two Bernstein surfaces being multiplied. 
\end{property}

\begin{property}[Derivatives]
The partial derivatives of a Bernstein surface can be calculated by multiplying a differentiation matrix with the surface's control points. For example, consider the Bernstein surface representing the rotation of the rod, Equation \ref{eq:phi}, and let \(\bar{\mathbf{\Phi}}^{m,n}\) be the matrix of control points, i.e., \(\{\bar{\mathbf{\Phi}}^{m,n}\}_{i,j} = \bar{\phi}_{i,j}^{m,n}\). The partial derivatives $\frac{\partial}{\partial s} \phi(s,t)$ and $\frac{\partial}{\partial t} \phi(s,t)$ are given by 
\begin{equation} \label{eq:bsurfderiv}
        \bar{\mathbf{\Phi}}^{m-1,n}=\textbf{D}_m^\top \bar{\mathbf{\Phi}}^{m,n}, \quad
        \bar{\mathbf{\Phi}}^{m,n-1}=\bar{\mathbf{\Phi}}^{m,n}\textbf{D}_n ,
\end{equation}
respectively, with
\begin{equation} \label{eq:bsurfderiv}
        \textbf{D}_n= \frac{n}{T}
        \begin{bmatrix}
            -1 & 0 & \ldots & 0 \\
            1 & \ddots & \ddots & \vdots \\
            0 & \ddots & \ddots & 0 \\
            \vdots & \ddots & \ddots & -1 \\
            0 & \ldots & 0 & 1 
        \end{bmatrix} \in \mathbb{R}^{n+1\times n}.
\end{equation}
\begin{equation} \label{eq:bsurfderiv}
        \textbf{D}_m= \frac{m}{L}
        \begin{bmatrix}
            -1 & 0 & \ldots & 0 \\
            1 & \ddots & \ddots & \vdots \\
            0 & \ddots & \ddots & 0 \\
            \vdots & \ddots & \ddots & -1 \\
            0 & \ldots & 0 & 1 
        \end{bmatrix} \in \mathbb{R}^{m+1\times m}.
\end{equation}
The operation performed for obtaining derivatives will lower the degree of the resulting polynomial by one. 
\end{property}

Using the two properties above, the expressions on the right hand side of Equation \eqref{eq:vconstraintsder} can also be represented as Bernstein polynomials:
\begin{equation} \label{eq:vexpressions}
\begin{aligned}
||\mathbf{p}_s||^2 & = \sum_{i=0}^{2m}\sum_{j=0}^{2n}\bar{{v}}_{i,j}^{2m,2n}B_{i}^{2m}(s)B_{j}^{2n}(t) \\ 
||\mathbf{p}_t||^2 & = 
\sum_{i=0}^{2m}\sum_{j=0}^{2n}\bar{{q}}_{i,j}^{2m,2n}B_{i}^{2m}(s)B_{j}^{2n}(t) \\
||\, [\phi_s ,  \theta_s, \psi_s] \, ||^2 & = 
\sum_{i=0}^{2m}\sum_{j=0}^{2n}\bar{{u}}_{i,j}^{2m,2n}B_{i}^{2m}(s)B_{j}^{2n}(t) \\
|| \, [\phi_t ,  \theta_t, \psi_t] \, ||^2 & = 
\sum_{i=0}^{2m}\sum_{j=0}^{2n}\bar{{\omega}}_{i,j}^{2m,2n}B_{i}^{2m}(s)B_{j}^{2n}(t) \\
||\mathbf{p}_{ss}||^2 & = \sum_{i=0}^{2m}\sum_{j=0}^{2n}\bar{{a}}_{i,j}^{2m,2n}B_{i}^{2m}(s)B_{j}^{2n}(t) \\ 
||\mathbf{p}_{tt}||^2 & = 
\sum_{i=0}^{2m}\sum_{j=0}^{2n}\bar{{\textfrak{a}}}_{i,j}^{2m,2n}B_{i}^{2m}(s)B_{j}^{2n}(t)
\end{aligned}
\end{equation}
The coefficients $\bar{{v}}_{i,j}^{2m,2n}$, $\bar{{q}}_{i,j}^{2m,2n}$, $\bar{{u}}_{i,j}^{2m,2n}$, and \(\bar{{\omega}}_{i,j}^{2m,2n}\) can be obtained from algebraic manipulation of the Bernstein coefficients of $\textbf{p}(s,t)$, \(\phi(s,t)\), \(\theta(s,t)\) and \(\psi(s,t)\).

\begin{figure}
    \centering
    \includegraphics[width = 0.3\textwidth]{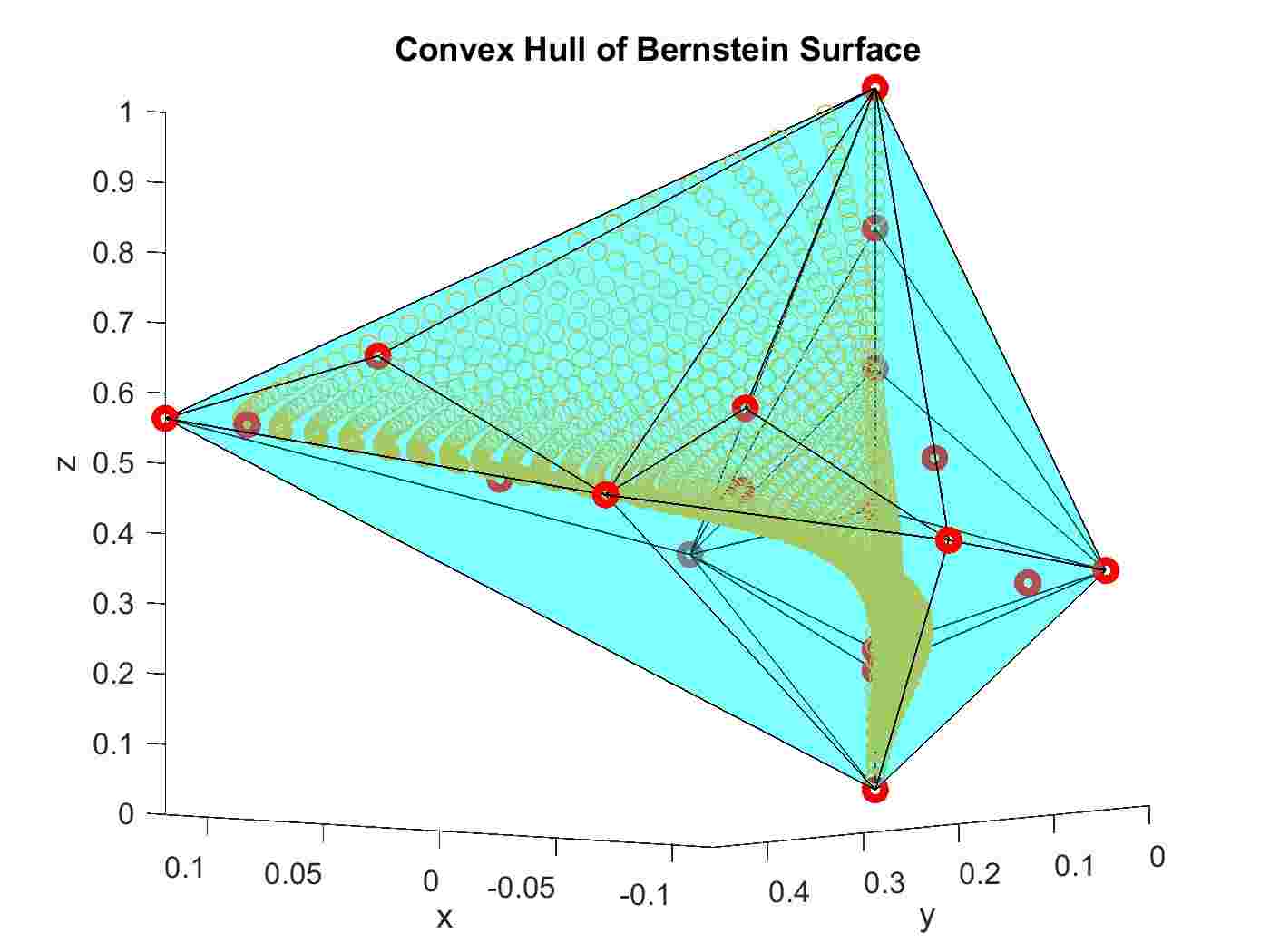}
    \caption{This is the convex hull of a Bernstein surface of order m=5, n=5. Here the red dots represent control points of the surface, the orange dots represent points on the surface, and the blue polygon is the convex hull of the surface.}
    \label{fig:convex_hull}
\end{figure}

To enforce the inequality constraints we use the convex hull property of Bernstein surfaces.
\begin{property}[Convex Hull]
A Bernstein surface lies within the convex hull defined by its control points. 
\end{property}
An illustration of this property is provided in Figure \ref{fig:convex_hull}.
For the surface defined in \eqref{eq:vexpressions}, for example, this means
\begin{equation}
   ||\mathbf{p}_s||^2 \leq \max_{i,j} \bar{{v}}_{i,j}^{2m,2n}
\end{equation}
for $i={0,...2m}$ and $j={0,...,2n}$.
Essentially, this property offers a mathematical tool for enforcing boundary constraints by directly imposing them onto the surface's control points. However, the convex hull's enclosure can be considerably more expansive than the actual curve it encompasses, possibly leading to inefficiencies. For a refined representation, one might consider leveraging the degree elevation property.

\begin{property}[Degree elevation] \label{prop:degelev}
For \({n_e}>n\) and \({m_e>m}\), the degree of a surface with matrix of control points  $\bar{\mathbf{G}}^{m,n} = \{\bar{g}_{i,j}^{m,n} \}$ can be degree elevated as follows
\begin{equation}
    \begin{split}
        \textbf{G}^{{m_e},n} = \textbf{E}_m^{{m_e}\top}\textbf{G}^{m,n}, \qquad
        \textbf{G}^{m,{n_e}} = \textbf{G}^{m,{n}}\textbf{E}_n^{n_e} ,
    \end{split}
\end{equation}
where \(\textbf{E}_n^{n_e}=\{e_{j,k}\}\in\mathbb{R}^{(n+1)\times({n_e}+1)} \)
\begin{equation} \label{eq:degel}
\begin{split}
        e_{i,i+j}= \left\{ \begin{matrix}
            \frac{{{n_e}-n \choose j}{n \choose i}}{{{n_e} \choose i+j}} & j\le {n_e}-n, \ i\le n \\ \\ 
            0 & \text{otherwise}
        \end{matrix} \right. 
\end{split}
\end{equation}
\end{property}
\begin{figure}
    \centering
    \includegraphics[width=0.3\textwidth]{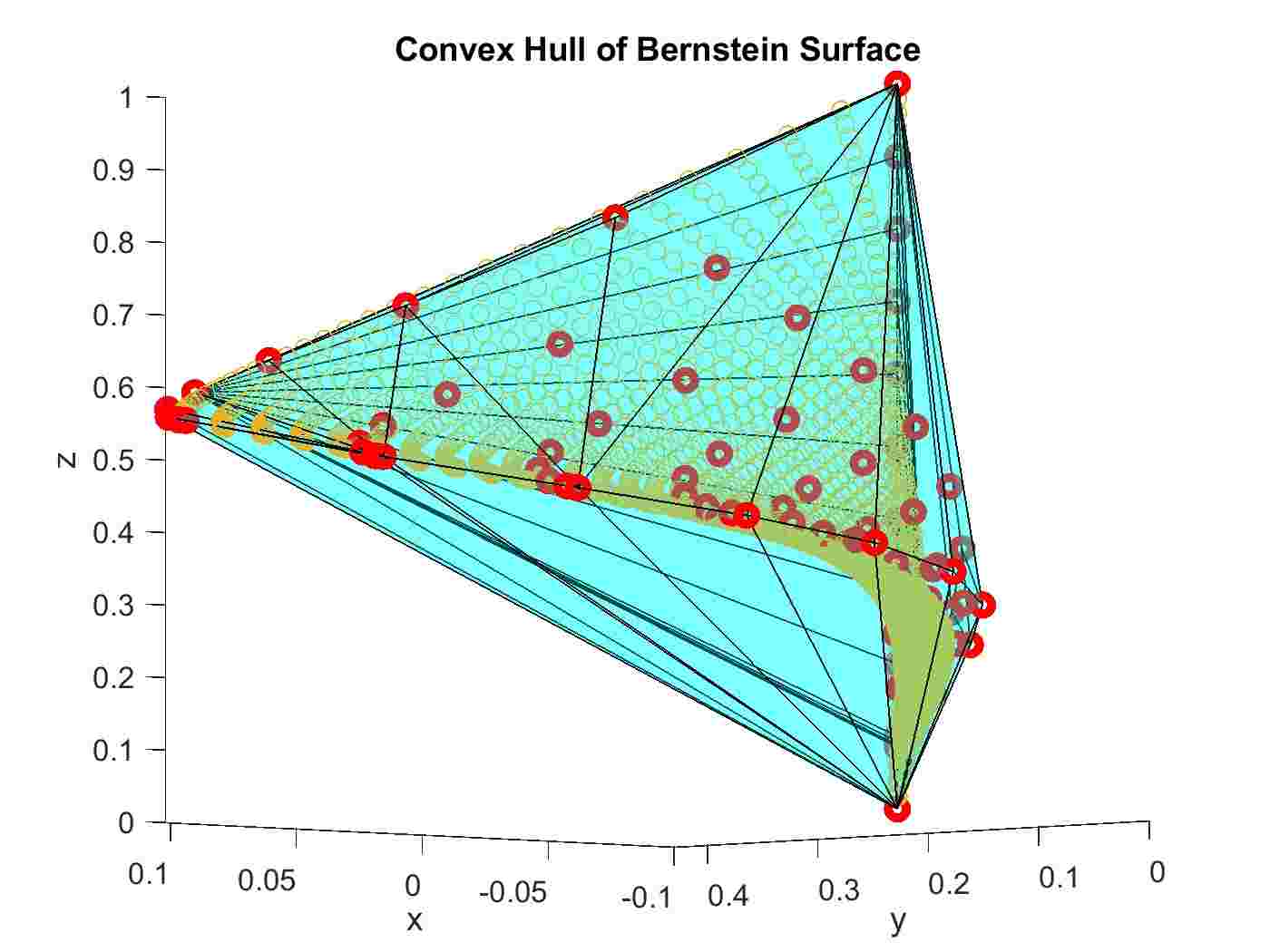}
    \caption{This is the sane surface in Figure \ref{fig:convex_hull} after being degree elevated to m=10, n=10.}
    \label{fig:deg_el}
\end{figure}
A graphical depiction of this property is provided in Figure \ref{fig:deg_el}.
When a Bernstein surface undergoes degree elevation, its control points converge to the surface. For theoretical insights on the convergence of these control points to the polynomial during degree elevation, refer to \cite{prautzsch1994convergence}.
With this is mind, we employ the degree elevation property to ensure dynamic feasibility and adherence to collision avoidance constraints. In particular, we degree elevate the Bernstein surfaces in Equation \eqref{eq:vexpressions} to obtain equivalent surfaces of degrees \(2m_e, 2n_e\). Then, constraints \eqref{eq:vconstraintsder} are imposed as follows
\begin{equation} \label{eq:vconstraintsdisc}
\begin{aligned}
&v_{\min}^2 \leq \bar{{v}}_{i,j}^{2m_e,2n_e} \leq v_{\max}^2, \quad &\bar{{q}}_{i,j}^{2m_e,2n_e} \leq q_{\max}^2, \\
&\bar{{u}}_{i,j}^{2m_e,2n_e}\leq u_{\max}^2, \quad &\bar{{\omega}}_{i,j}^{2m_e,2n_e}\leq \omega_{\max}^2, \\
& \bar{{a}}_{i,j}^{2m_e,2n_e} \leq v_{s,\max}^2, \quad &\bar{{\textfrak{a}}}_{i,j}^{2m_e,2n_e} \leq q_{t,\max}^2, 
\end{aligned}
\end{equation}
for all \(i=0,\ldots,2m_e\), \(j=0,\ldots,2n_e\).

For what concerns the boundary conditions, the endpoint value property of Bernstein surfaces is of relevance.

\begin{property}[End point values]
Terminal points of Bernstein surfaces fall on their corresponding control points, e.g., for the Bernstein surface in Equation \eqref{eq:BSposition} we have
\begin{equation}
\begin{split}
    \textbf{p}(0,0) = \bar{\textbf{p}}^{m,n}_{0,0}, \quad \textbf{p}(0,T) = \bar{\textbf{p}}^{m,n}_{0,n},    \\
    \textbf{p}(L,0) = \bar{\textbf{p}}^{m,n}_{m,0}, \quad \textbf{p}(L,T) = \bar{\textbf{p}}^{m,n}_{m,n} .    \\
\end{split}
\end{equation}
Further, edges of the surface are given by the Bernstein polynomial of the edge's control points, e.g.
\begin{equation}
    \textbf{p}(s,0) = \sum_{i=0}^m \bar{\textbf{p}}^{m,n}_{i,0} B_{i}^{m}(s).
\end{equation}
\end{property}

Then, using the above property, the boundary conditions in Equation \eqref{eq:boundary} can be expressed as follows
\begin{equation} \label{eq:boundarydisc} 
\begin{aligned} 
\tilde{\mathbf{g}}_i(\bar{\textbf{p}}^{m,n}_{i,0},\bar{\phi}^{m,n}_{i,0},\bar{\theta}^{m,n}_{i,0},\bar{\psi}^{m,n}_{i,0}) & = 0 \\
\tilde{\mathbf{g}}_f(\bar{\textbf{p}}^{m,n}_{i,n},\bar{\phi}^{m,n}_{i,n},\bar{\theta}^{m,n}_{i,n},\bar{\psi}^{m,n}_{i,n}) & = 0 \\
\tilde{\mathbf{g}}_0(\bar{\textbf{p}}^{m,n}_{0,j},\bar{\phi}^{m,n}_{0,j},\bar{\theta}^{m,n}_{0,j},\bar{\psi}^{m,n}_{0,j}) & = 0, 
\end{aligned}
\end{equation} 
for \(i=0,\ldots,m\) and \(j=0,\ldots,n\), where the functions \(\tilde{\mathbf{g}}_i()\), \(\tilde{\mathbf{g}}_f()\) and \(\tilde{\mathbf{g}}_0()\) are obtained by expressing the functionals \({\mathbf{g}}_i()\), \({\mathbf{g}}_f()\) and \({\mathbf{g}}_0()\) in terms of the control points of the Bernstein surfaces and their derivatives. 

The de Casteljau Algorithm is introduced next to deal with the obstacle avoidance constraints.

\begin{figure}
    \centering
    \includegraphics[width = 0.3\textwidth]{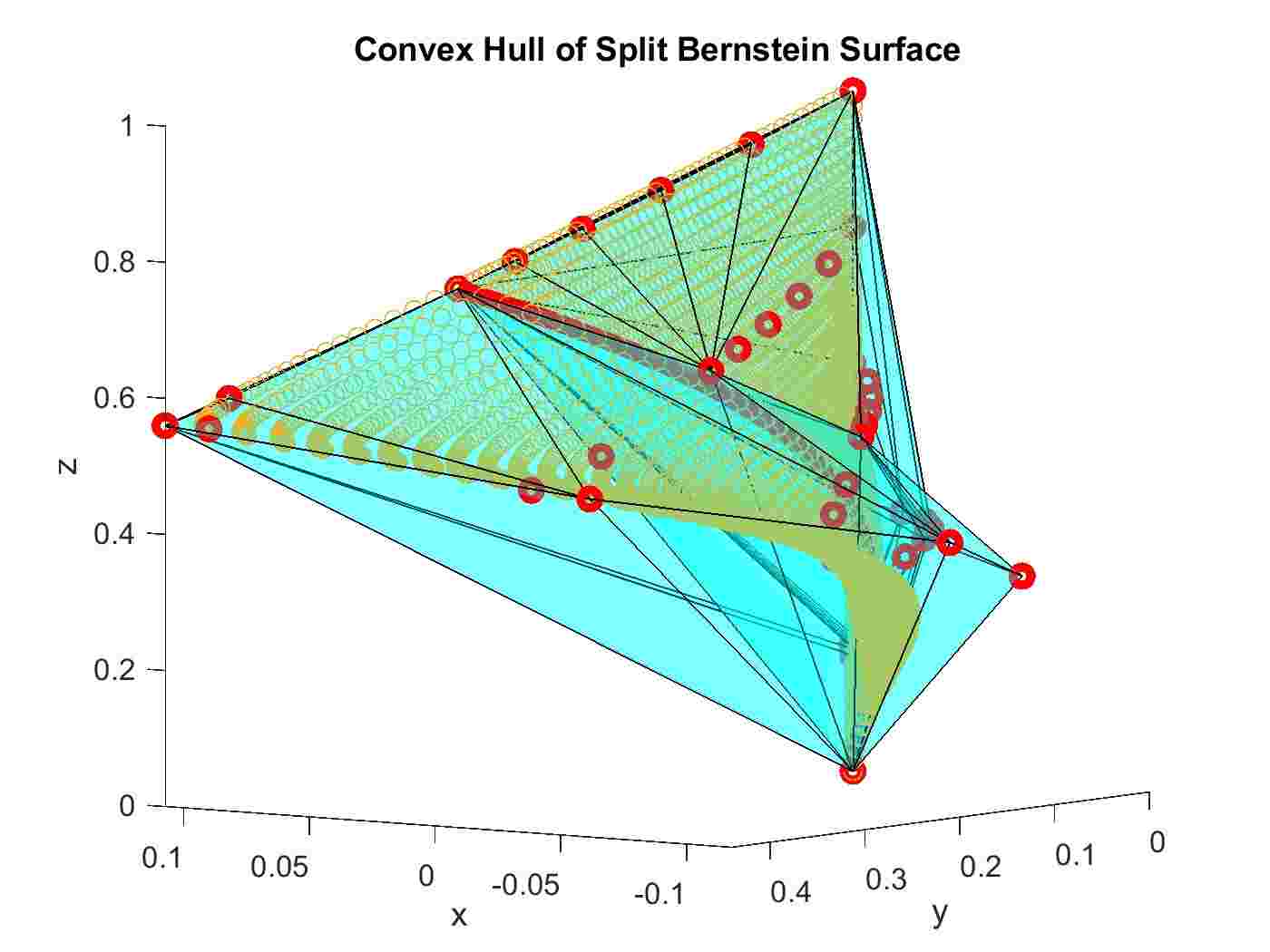}
    \caption{Here, the surface seen in Figure \ref{fig:convex_hull} is split once along the plotted red line using the de Casteljau algorithm.}
    \label{fig:deCastel}
\end{figure}

\textit{The de Casteljau Algorithm:} Using the de Casteljau algorithm, a Bernstein surface can be partitioned into two distinct surfaces. For a Bernstein surface of order \(m \times n\), as depicted in Equation \eqref{eq:BSposition}, when provided with a scalar \(s_{\text{div}} \in [0, L]\) or \(t_{\text{div}} \in [0, T]\), the algorithm determines the Bernstein polynomial at the chosen division point and generates two subsequent Bernstein surfaces from the subdivision.
As an illustration, for \(s_{\text{div}} \in [0, L]\), the algorithm produces control points of Bernstein polynomial \(\mathbf{p}(s_{\text{div}}, t)\) and control points of Bernstein surfaces \(\mathbf{p}_1\) and \(\mathbf{p}_2\) such that:
\[
\mathbf{p}(s, t) = 
\begin{cases} 
\mathbf{p}_1\left(\frac{s}{s_{\text{div}}}, t\right) & \text{for } s \in [0, s_{\text{div}}] \\
\mathbf{p}_2\left(\frac{s - s_{\text{div}}}{L - s_{\text{div}}}, t\right) & \text{for } s \in [s_{\text{div}}, 1]
\end{cases} \, \quad t \in [0,1] .
\]
An illustration of this algorithm is provided in Figure \ref{fig:deCastel}.

To determine the minimum separation between a Bernstein surface and a convex shape, i.e., the left hand side of Equation \eqref{eq:collisionavoidance}, we extend the technique initially proposed in \cite{chang2011computation} and implemented in \cite{kielas2022bernstein}. The approach incorporates the Convex Hull property, the End Point Values property, the de Casteljau Algorithm, and the Gilbert-Johnson-Keerthi (GJK) algorithm \cite{gilbert1988fast}. The GJK algorithm is a fast algorithm for determining minimum separations between convex objects.

The computation procedure is captured in Algorithm \ref{algorithm:spatSep}. The primary inputs to this function are the sets of Bernstein coefficients, denoted as $\bar{\textbf{P}}^{m,n}$, defining the Bernstein surface representing the position of the rod, and the vertices of the convex shape that must be avoided, $\bar{\textbf{Q}}$. Other inputs include the global upper bound on the minimum separation, $\alpha$, and a desired tolerance, $\epsilon$.

The function `upper\_bound()` computes the maximum separation between the end points of the Bernstein surface and the convex shape. This function is rooted in the End Point Values.
Conversely, the function `lower\_bound()` determines the lower bound of the separation between the surface and the shape using the GJK algorithm, backed by the Convex Hull property. The algorithm subsequently checks for specific early termination conditions. If met, it returns the current estimate. If not, the Bernstein surface and the convex shape are subdivided using the de Casteljau algorithm. The process then recursively determines the minimum distance for each subdivided segment, updating the threshold value as needed. The function ultimately provides a refined minimum distance estimate as its result.

\begin{algorithm}
        \caption{\(\text{MinDist}(\bar{\textbf{P}}^{m,n},\bar{\textbf{Q}},\alpha,\epsilon)\)}
        \label{algorithm:spatSep}

        $upper =$ upper\_bound($\bar{\textbf{P}}^{m,n},\bar{\textbf{Q}}$) \label{line:spat:ub}

        $lower =$ lower\_bound($\bar{\textbf{P}}^{m,n},\bar{\textbf{Q}}$) \label{line:spat:lb}

        \If{$upper < \alpha$}{ \label{line:spat:checkalpha}

            $\alpha = upper$
        }

        \If{$\alpha < lower$}{ \label{line:spat:prune}
            \Return $\alpha$
        }

        \eIf{$upper - lower < \epsilon$}{ \label{line:spat:checktol}
            \Return $\alpha$
        }{

            $\bar{\textbf{P}}_A^{m,n},\bar{\textbf{P}}_B^{m,n} =$ deCast($\bar{\textbf{P}}^{m,n}$)

            $\mathbf{Q}_A, \mathbf{Q}_B =$ deCast($\mathbf{Q}$)

            $\alpha =$ min($\alpha$,
            Algorithm \ref{algorithm:spatSep}
            ($\mathbf{P}_A^{m,n}, \mathbf{Q}_A, \alpha$))

            $\alpha =$ min($\alpha$,
            Algorithm \ref{algorithm:spatSep}
            ($\mathbf{P}_A^{m,n}, \mathbf{Q}_B, \alpha$))

            $\alpha =$ min($\alpha$,
            Algorithm \ref{algorithm:spatSep}
            ($\mathbf{P}_B^{m,n}, \mathbf{Q}_A, \alpha$))

            $\alpha =$ min($\alpha$,
            Algorithm \ref{algorithm:spatSep}
            ($\mathbf{P}_B^{m,n}, \mathbf{Q}_B, \alpha$))
        }
        \Return $\alpha$
    \end{algorithm}

\begin{remark} \label{rem:minsep}
   Algorithm \ref{algorithm:spatSep} can be adapted to calculate the minimum separation among several Bernstein surfaces. This is relevant for multi-robot systems where multiple rods collaborate in near proximity, such as in multi-tentacle systems. Nonetheless, this topic extends beyond the purview of this paper.
\end{remark}

Considering Algorithm \ref{algorithm:spatSep}, the obstacle avoidance constraint is formulated as:
\begin{equation} \label{eq:collisionavoidancedisc}
    \text{MinDist}(\bar{\textbf{P}}^{m,n},\bar{\textbf{Q}},\alpha,\epsilon) \geq d_{\text{safe}} > 0 .
\end{equation}

Finally, for the approximation of the cost, let us define
\begin{equation*}  
\tilde{\ell} (\mathbf{p}(s,t),\phi(s,t),\theta(s,t),\psi(s,t)) \triangleq \ell (\mathbf{p}(s,t),\mathbf{R}(s,t))),
\end{equation*}
i.e., \(\tilde{\ell}()\) is obtained by expressing \(\ell()\) in terms of the position and Euler angles. Then, the approximation of Problem \ref{prob:continuous} can be expressed as follows. 

\vspace{5px}

\begin{prob} \label{prob:discrete}
Find \(\bar{\mathbf{p}}^{m,n}_{i,j},\bar{\phi}^{m,n}_{i,j},\bar{\theta}^{m,n}_{i,j},\bar{\psi}^{m,n}_{i,j},T\) that minimize
\begin{equation}
 \sum_{i=0}^{m}\sum_{j=0}^{n} \frac{L}{m+1}\frac{T}{n+1} \tilde{\ell} (\bar{\mathbf{p}}^{m,n}_{i,j},\bar{\phi}^{m,n}_{i,j},\bar{\theta}^{m,n}_{i,j},\bar{\psi}^{m,n}_{i,j}) 
\end{equation}
subject to Equations \eqref{eq:vconstraintsdisc}, \eqref{eq:boundarydisc}, and \eqref{eq:collisionavoidancedisc}.
\end{prob}

\vspace{5px}

The solution to Problem \ref{prob:discrete} provides a set of optimal control points $\bar{\mathbf{p}}^{m,n}_{i,j},\bar{\phi}^{m,n}_{i,j},\bar{\theta}^{m,n}_{i,j},$ and $\bar{\psi}^{m,n}_{i,j}$, together with the time $T$. Using these coefficients, the respective Bernstein surfaces can be defined as in 
\begin{equation}
\mathbf{p}^*(s,t) = \sum_{i=0}^{m}\sum_{j=0}^{n} \bar{\mathbf{p}}^{m,n}_{i,j} B_i^m(s) B_j^n(t),
\end{equation}
with similar representations for $\phi^*_{i,j}(s,t)$, $\theta^*_{i,j}(s,t)$, and $\psi^*_{i,j}(s,t)$. 

From a theoretical standpoint, the following results hold regarding the feasibility and consistency of the proposed approximation method.
\begin{lemma}[Feasibility]
If Problem \ref{prob:continuous} is feasible, then there exist threshold orders \({m}^*\) and \({n}^*\), and for any approximation orders \(m \geq {m}^*\) and \(n \geq {n}^*\), there exist control points $\bar{\mathbf{p}}^{m,n}_{i,j},\bar{\phi}^{m,n}_{i,j},\bar{\theta}^{m,n}_{i,j},$ and $\bar{\psi}^{m,n}_{i,j}$ that constitute a feasible solution to Problem \ref{prob:discrete}.
\end{lemma}

\begin{lemma}[Consistency]
Assume that the cost function, \(\tilde{\ell}\) is Lipschitz continuous in its arguments. Then, as \(m,n \to \infty\), the solution to Problem \ref{prob:discrete}, $\bar{\mathbf{p}}^{m,n}_{i,j},\bar{\phi}^{m,n}_{i,j},\bar{\theta}^{m,n}_{i,j},$ and $\bar{\psi}^{m,n}_{i,j}$, gives Bernstein surfaces \(
\mathbf{p}^*(s,t) \), $\phi^*_{i,j}(s,t)$, $\theta^*_{i,j}(s,t)$, and $\psi^*_{i,j}(s,t)$ that are optimal solutions to Problem \eqref{prob:discrete}.  
\end{lemma}

The results above state that solutions to the approximated Problem \ref{prob:discrete} exist and converge to the optimal solution of the original Problem \ref{prob:continuous}. Due to space restrictions, detailed proofs are not provided here, but they are fundamentally grounded in the principles of the consistency approximation theory \cite{polak2012optimization}. The outline of these proofs aligns with the proofs presented in \cite{cichella2020optimal}. However, it is noteworthy to mention that while \cite{cichella2020optimal} primarily focuses on optimal control problems over ordinary differential equations (ODEs) using Bernstein polynomials, our current work broadens this scope to cover PDEs through the application of Bernstein surfaces.

\section{Numerical Results} \label{sec:res}

\begin{table}[]
\caption{Example Cases}
\label{table:case}
\centering
\begin{tabular}{|c||c|c|c|} 
    \hline
    & Case 1 & Case 2 & Case 3 \\
    \hline
    Init Pose & Straight & Curved & Straight\\
    Obstacles & 0 & 2 & 3\\
    $p_{des}$ & $\begin{bmatrix} 0.1 \\0.425 \\0.55 \end{bmatrix}$ & $\begin{bmatrix} 0.0 \\0.325 \\0.3 \end{bmatrix}$ & $\begin{bmatrix} 0.05 \\0.375 \\0.475 \end{bmatrix}$\\
    $\phi_{des},\theta_{des}$ & $-\pi/4,\pi/4$ & $-3\pi/4,\pi/8$ & $-\pi/2,3\pi/8$\\
    $\psi_{des}$ & $3\pi/4$ & 0 & $\pi/2$\\
    $q_{max}$ & 0.25 $m/s$ & 0.25 $m/s$ & 0.25 $m/s$\\
    $v_{max}$ & 1.15 & 1.25 & 1.25\\
    $v_{min}$ & 0.85 & 0.75 & 0.75\\
    $\omega_{max}$ & $\pi/4$ $s^{-1}$ & $\pi/2$ $s^{-1}$ & $\pi/2$ $s^{-1}$\\
    $u_{max}$ & $2\pi$ $m^{-1}$ & $2.5\pi$ $m^{-1}$ & $2.5\pi$ $m^{-1}$\\
    $v_{s,max}$ & 2 $m^{-1}$ & 3.25 $m^{-1}$ &  3.25 $m^{-1}$\\
    $q_{t,max}$ & 0.075 $m/s^2$ & 0.075 $m/s^2$ & 0.075 $m/s^2$\\
    $w_1$ & 100000 & 10000 & 10000\\
    $w_2=w_3$ & 1000 & 100 & 100\\
    $w_4$ & 100 & - & 100\\
    \hline
\end{tabular}
\end{table}

We validate the approach through three numerical scenarios. The cost function for all scenarios is given by
\[ 
\begin{split}
& \tilde{\ell}() =  \int_0^T \left( w_1 \Vert \textbf{p}(L,t)-\textbf{p}_{des} \Vert^2_2 + w_2 (\phi(L,t)-\phi_{des})^2 + \right. \\
& \qquad \left. w_3({\theta}(L,t))-\theta_{des})^2 ] + w_4 ({\psi}(L,t))-\psi_{des})^2 dt \right) \, ,
\end{split}
\]
with the corresponding constants enumerated in Table \ref{table:case}. This function seeks to optimize the rod's pose such that its end point reaches the desired destination and orientation in the least amount of time. Boundary conditions for each scenario are detailed in Table \ref{table:case}, as well as the dynamic constraints imposed into the rod. Initial conditions for all cases also include
\begin{equation}
\label{eq:IC}
    \begin{split}
        \textbf{q}(0,t) = \boldsymbol{\omega}(0,t) = [0 \quad 0 \quad 0]^\top, \\
        \textbf{p}(0,t)=[0 \quad 0 \quad 0]^\top, \\        
        \textbf{R}(0,t)=\textbf{I}_{3\times 3}.
    \end{split}
\end{equation}

\begin{figure}
    \centering
    \begin{subfigure}{.175\textwidth}
        \centering
        \includegraphics[width=\linewidth]{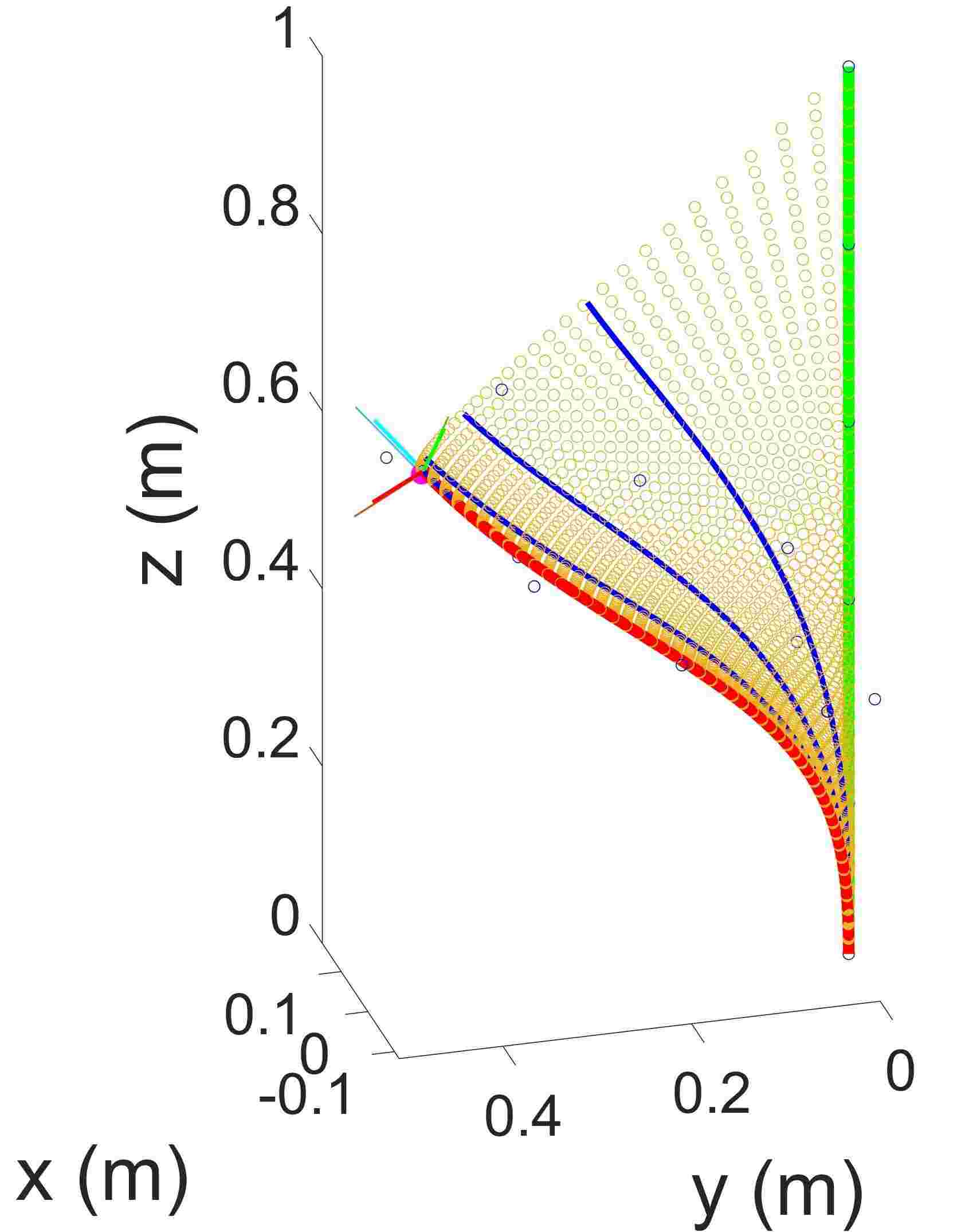}
        \caption{}
    \end{subfigure}
    \begin{subfigure}{.21\textwidth}
        \centering
        \includegraphics[width=\linewidth]{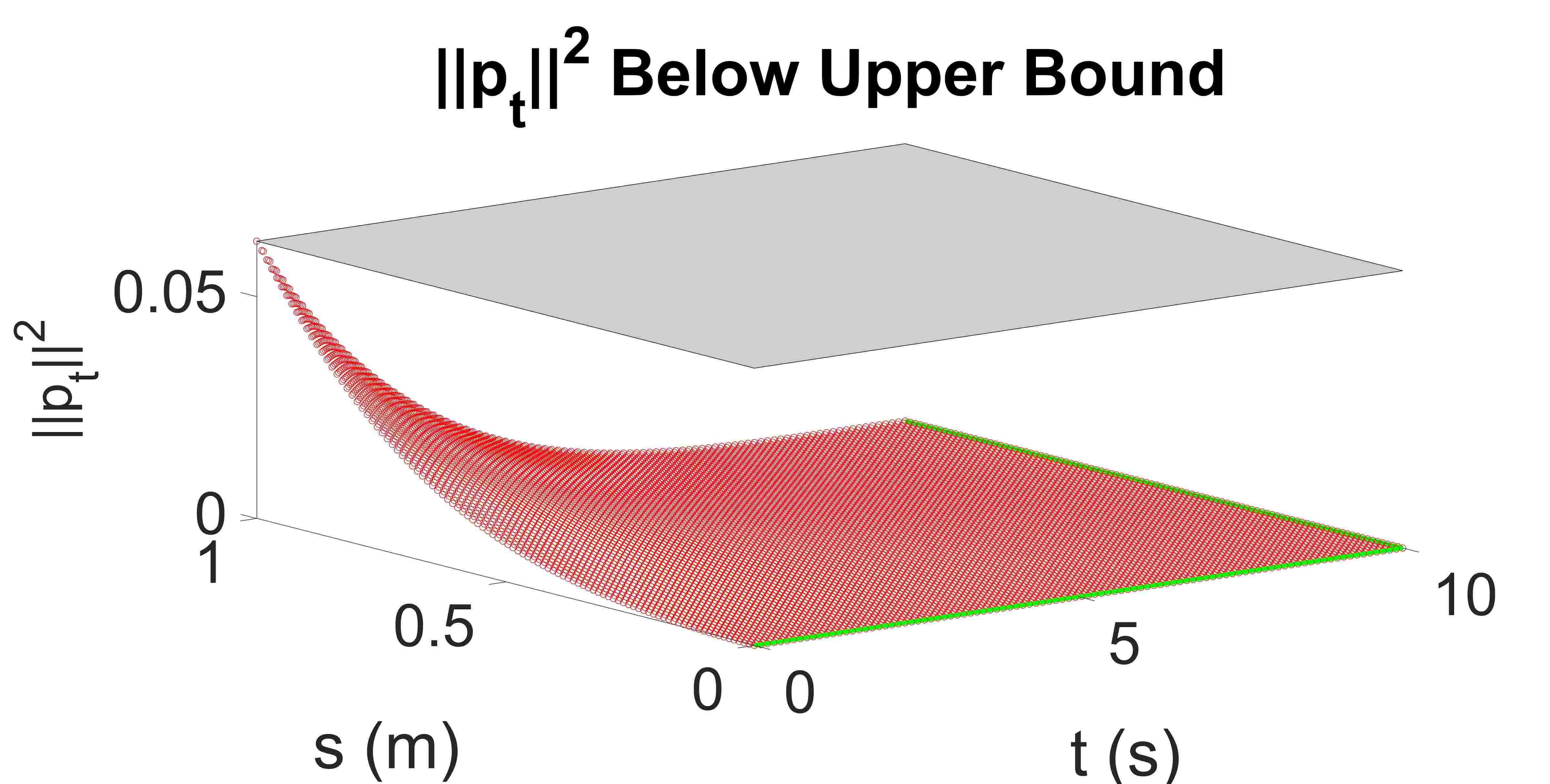}
        \caption{}
    \end{subfigure}
    \begin{subfigure}{.21\textwidth}
        \centering
        \includegraphics[width=\linewidth]{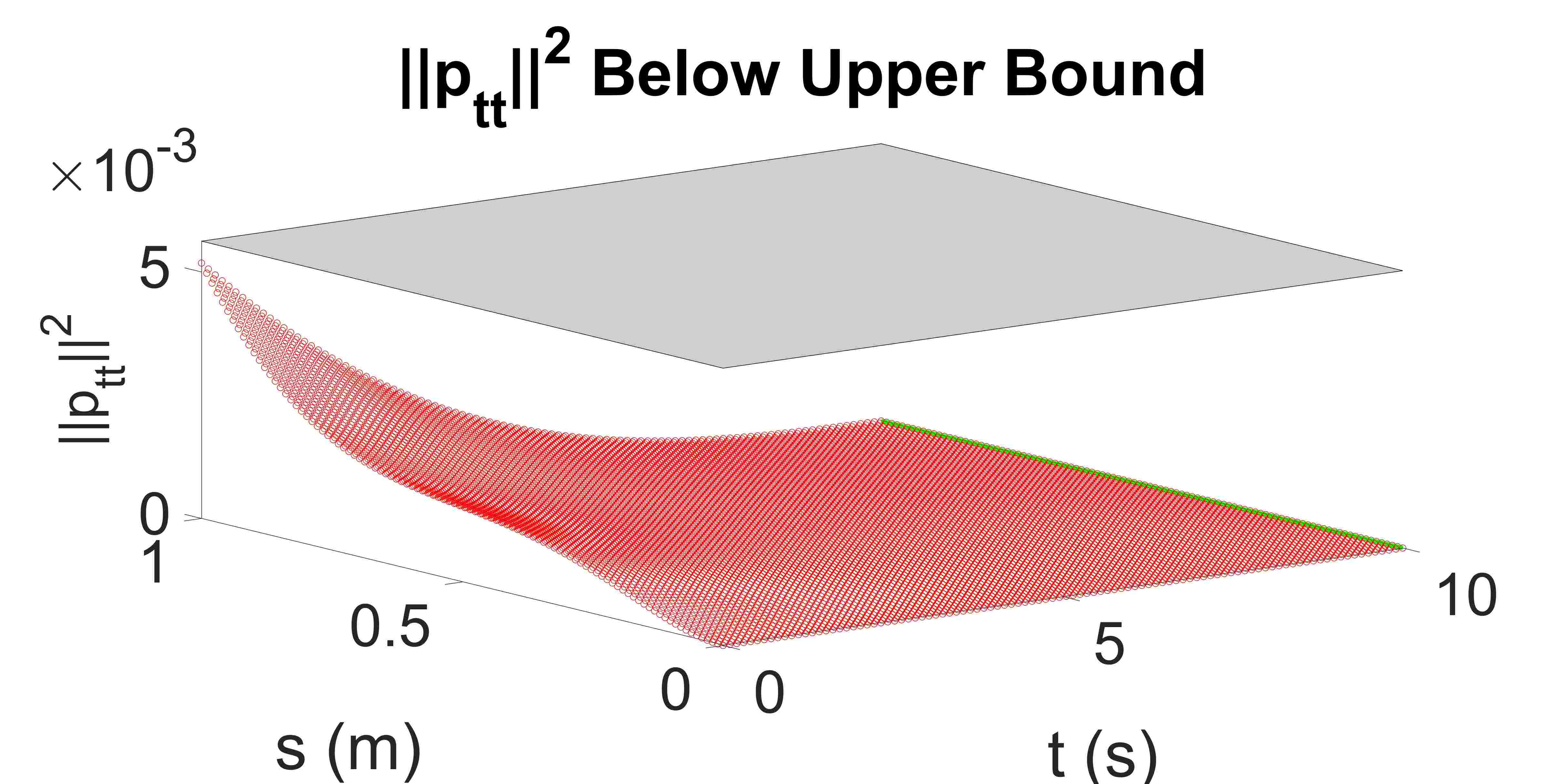}
        \caption{}
    \end{subfigure}
    \begin{subfigure}{.21\textwidth}
        \centering
        \includegraphics[width=\linewidth]{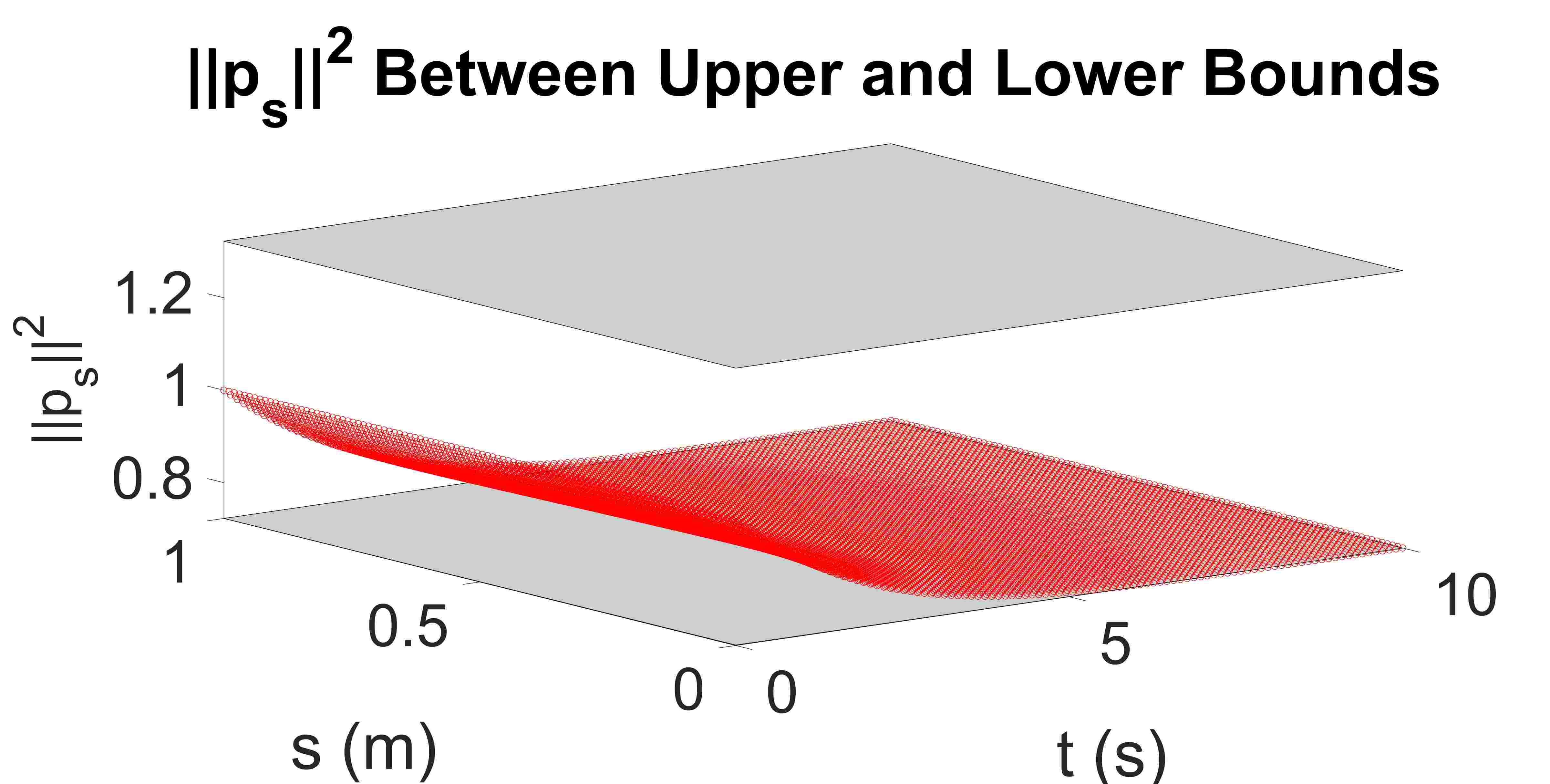}
        \caption{}
    \end{subfigure}
    \begin{subfigure}{.21\textwidth}
        \centering
        \includegraphics[width=\linewidth]{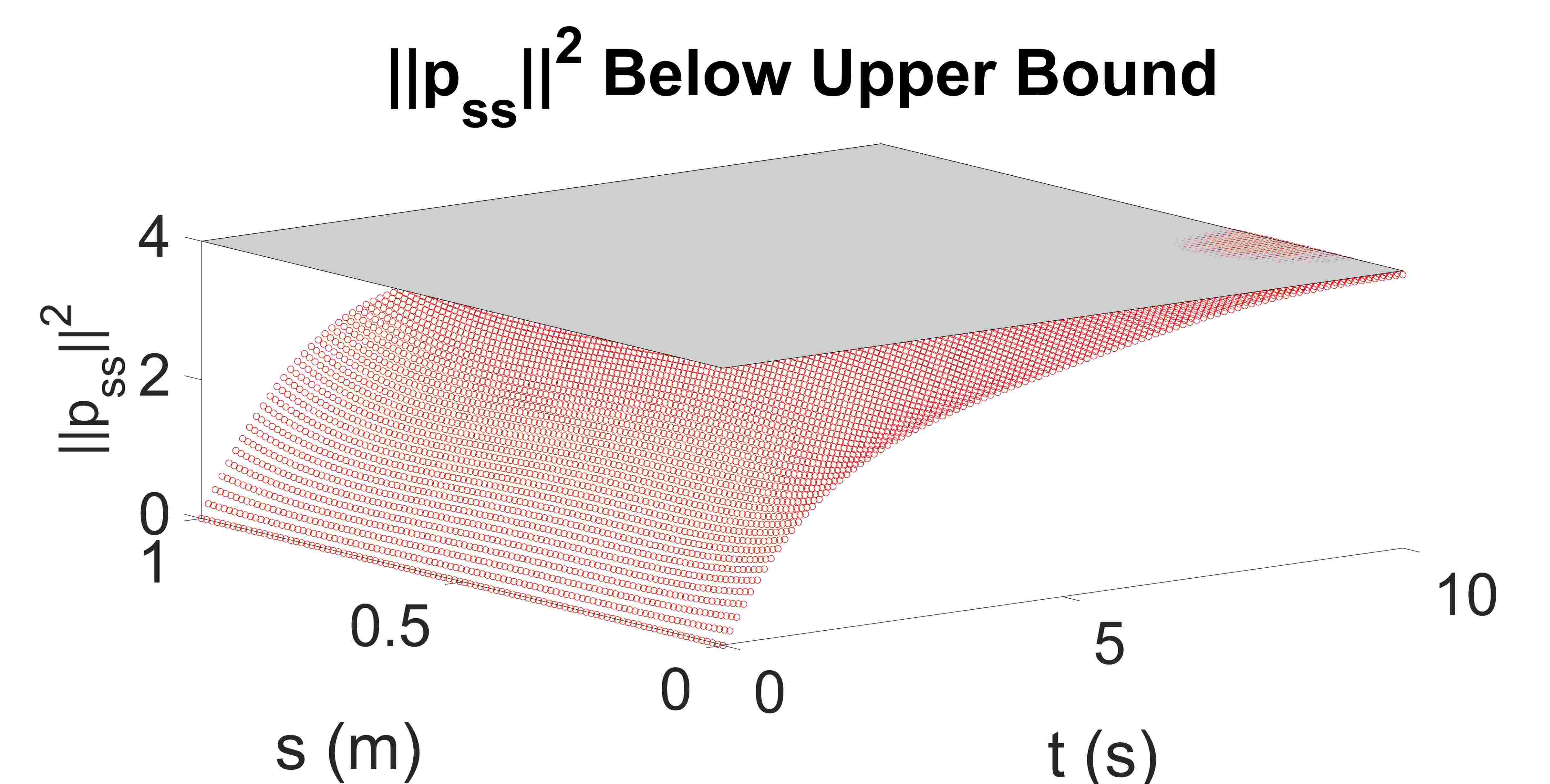}
        \caption{}
    \end{subfigure}
    \begin{subfigure}{.21\textwidth}
        \centering
        \includegraphics[width=\linewidth]{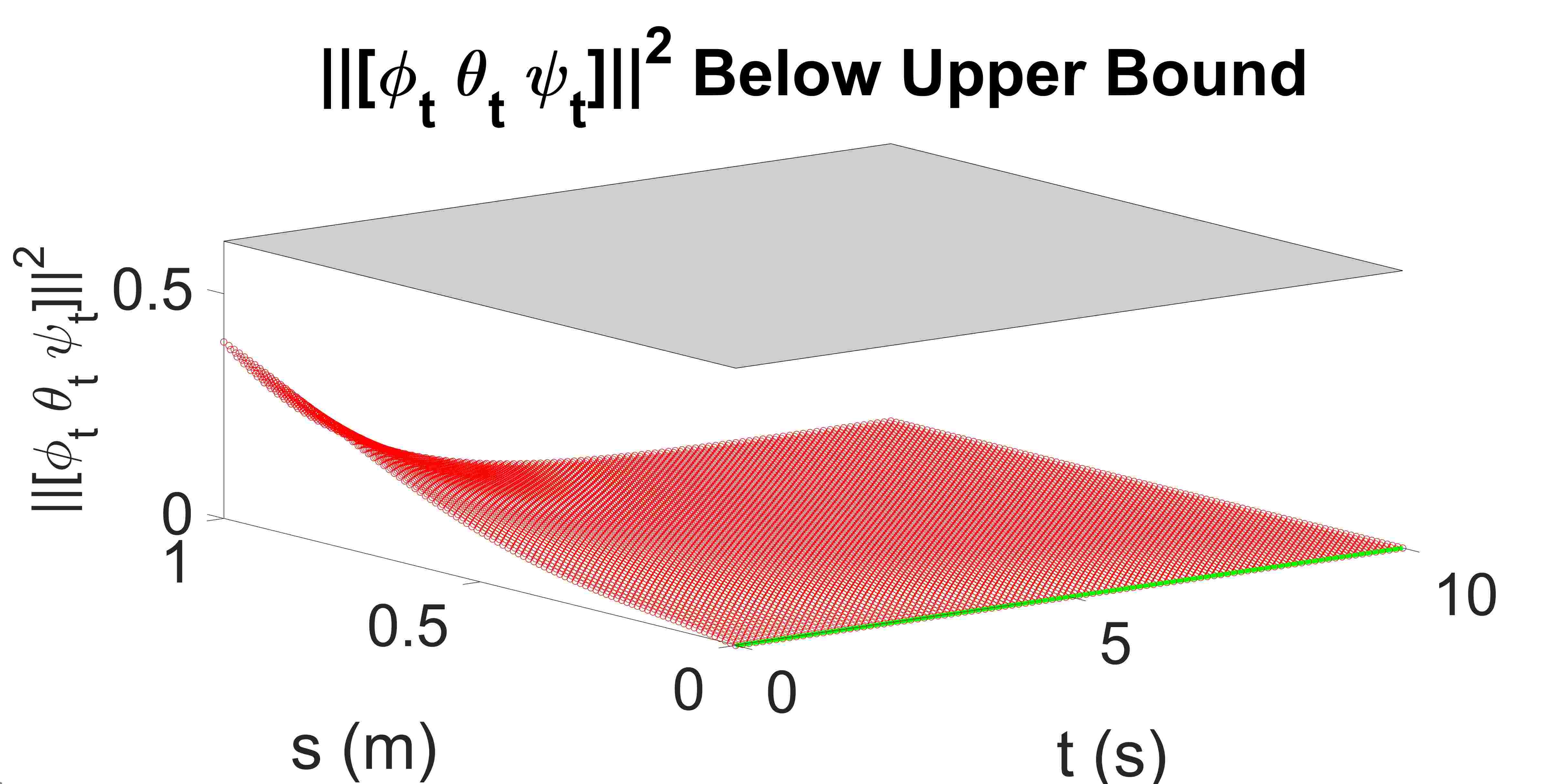}
        \caption{}
    \end{subfigure}
    \begin{subfigure}{.21\textwidth}
        \centering
        \includegraphics[width=\linewidth]{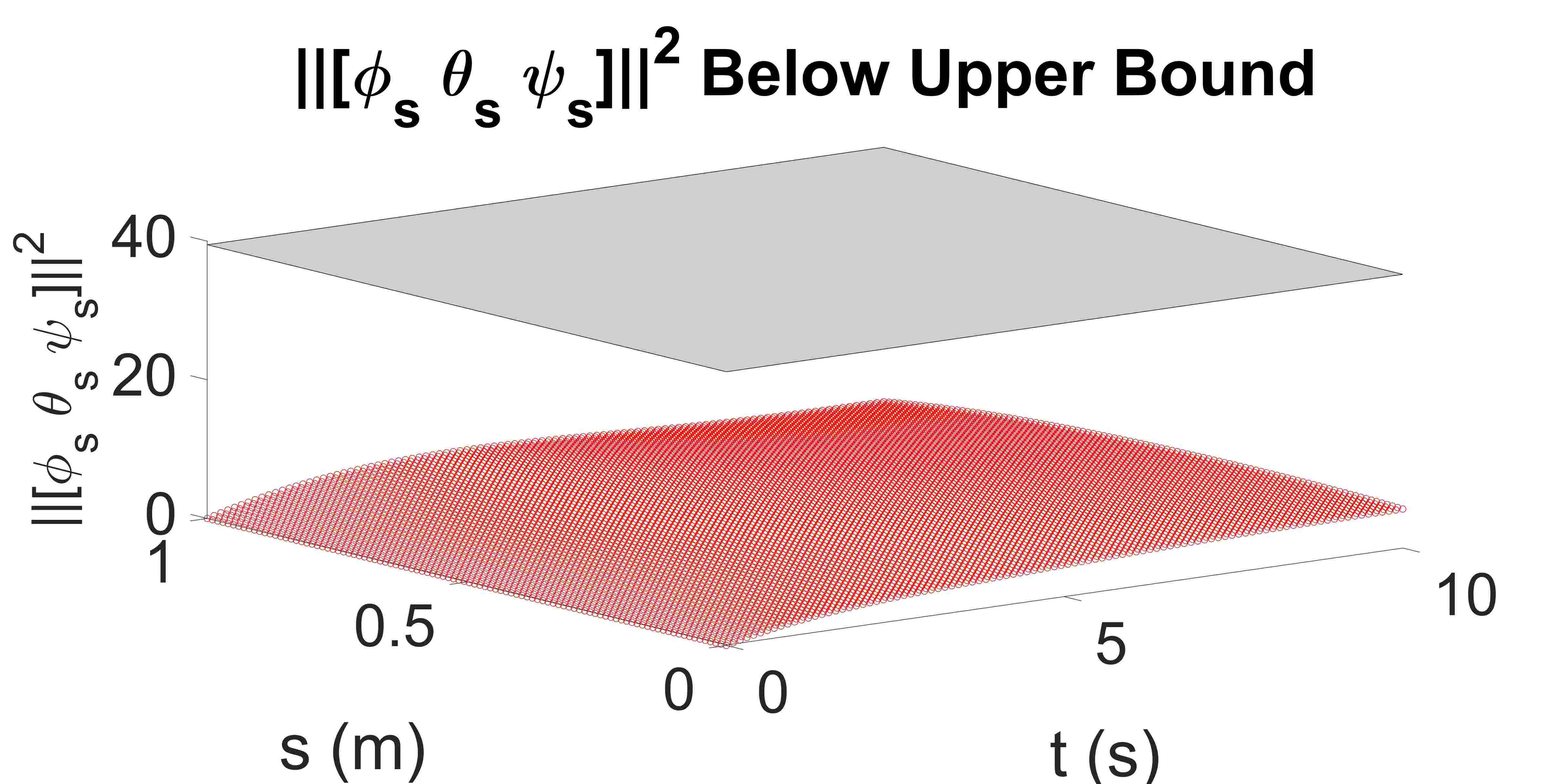}
        \caption{}
    \end{subfigure}
    \caption{Case 1: no obstacle planning.}
    \label{fig:case1}
\end{figure}

\begin{figure}
    \centering
    \begin{subfigure}{.25\textwidth}
        \centering
        \includegraphics[width=\linewidth]{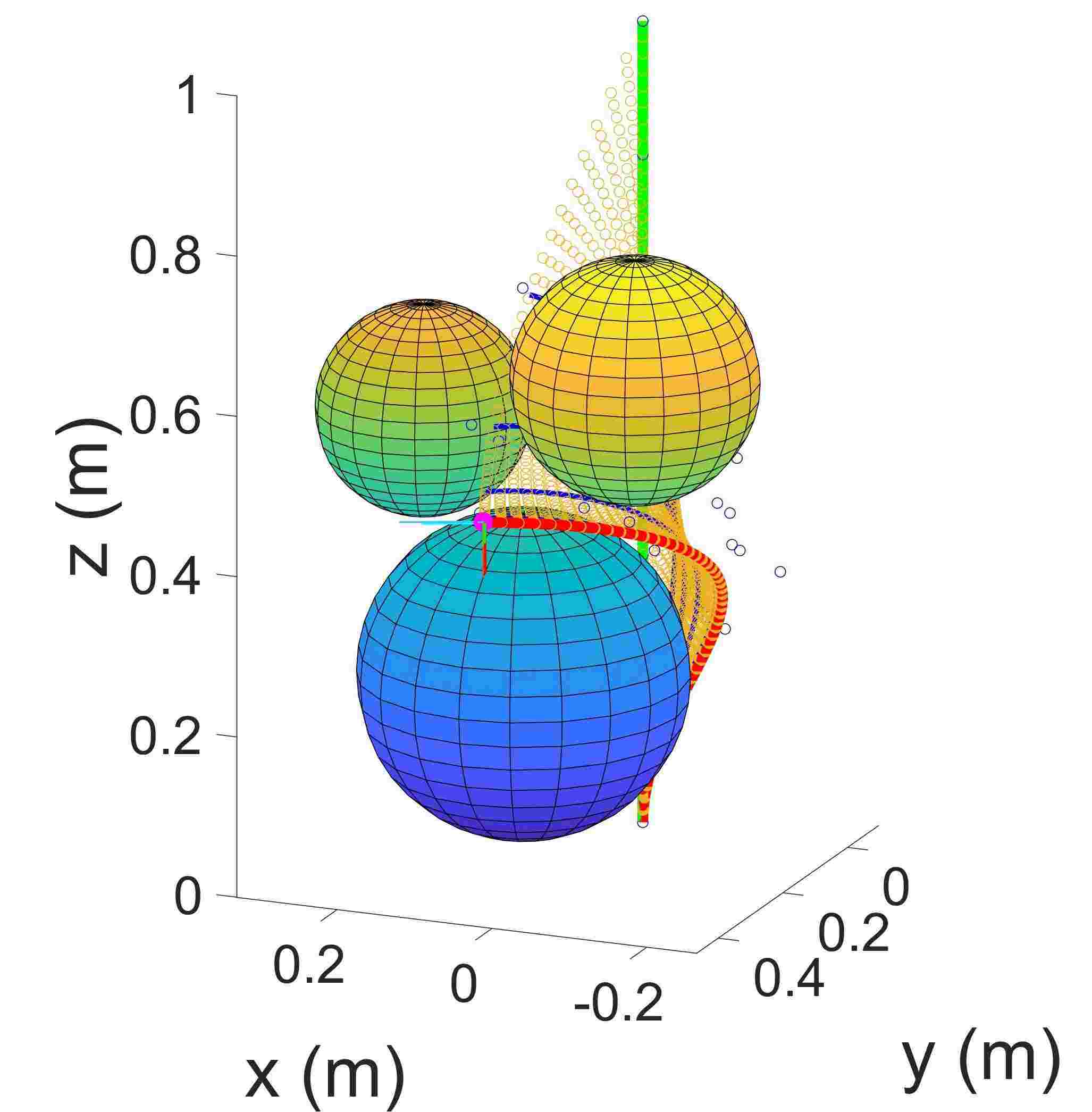}
        \caption{}
    \end{subfigure}
    \begin{subfigure}{.21\textwidth}
        \centering
        \includegraphics[width=\linewidth]{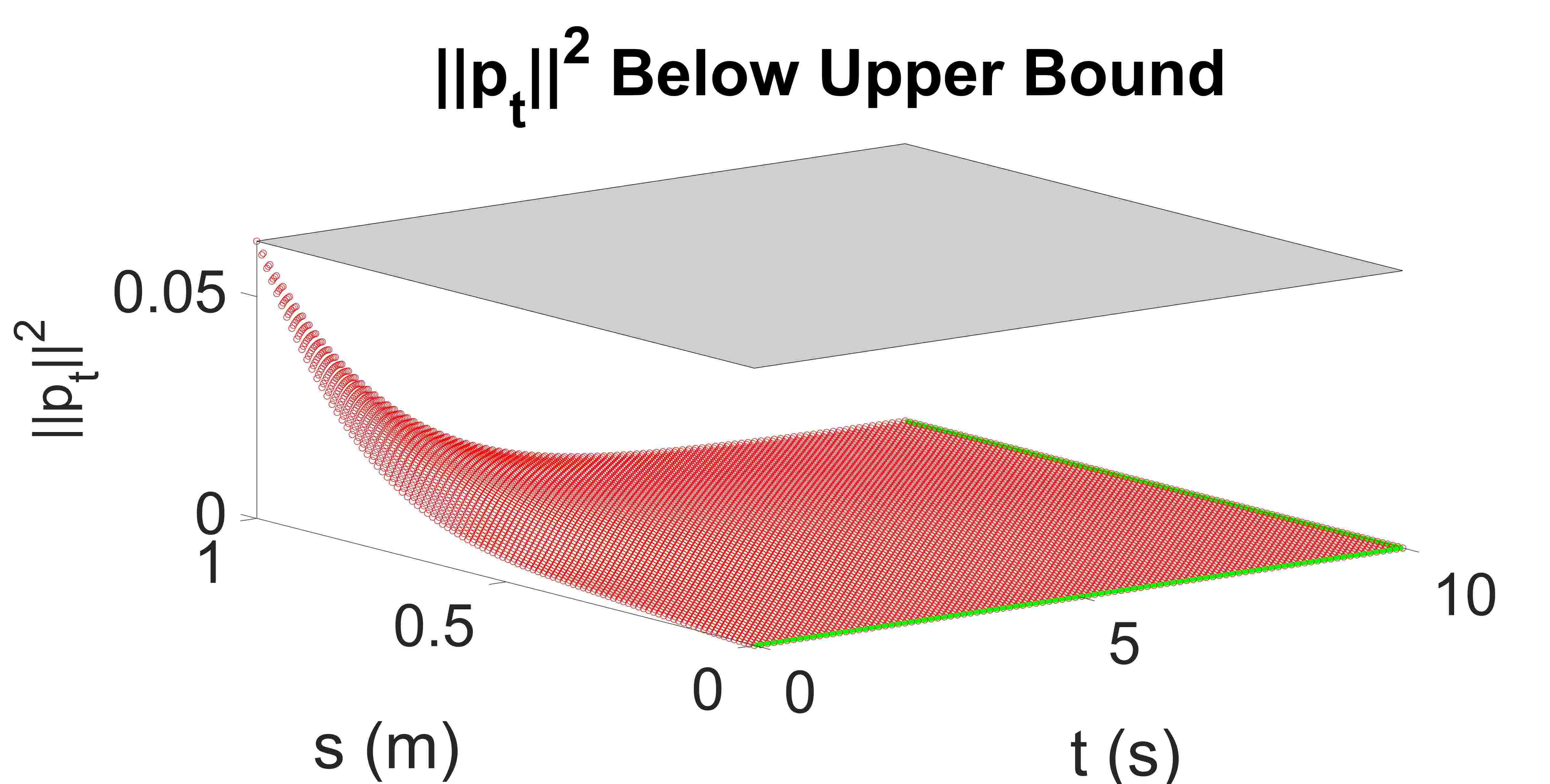}
        \caption{}
    \end{subfigure}
    \begin{subfigure}{.21\textwidth}
        \centering
        \includegraphics[width=\linewidth]{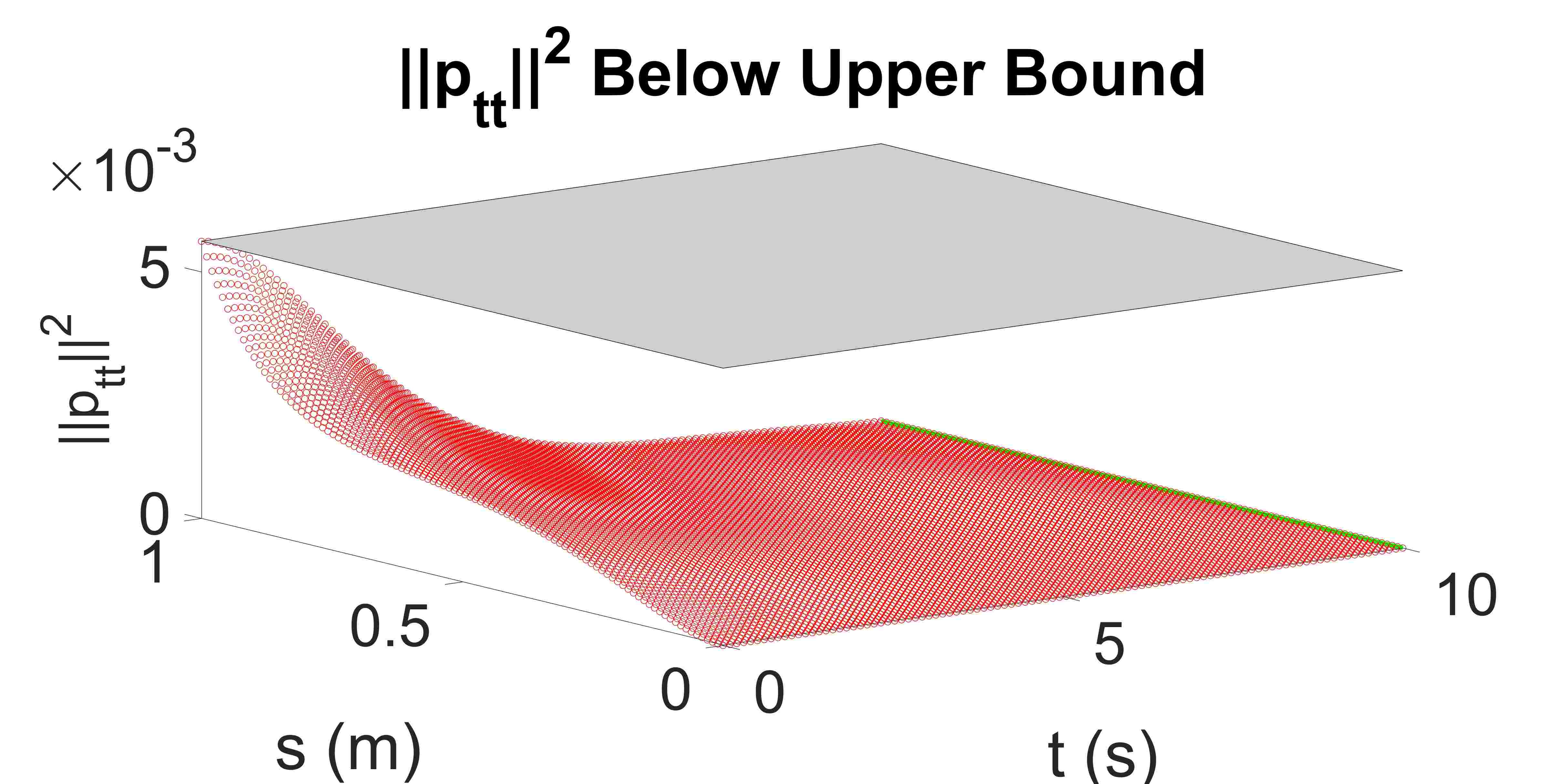}
        \caption{}
    \end{subfigure}
    \begin{subfigure}{.21\textwidth}
        \centering
        \includegraphics[width=\linewidth]{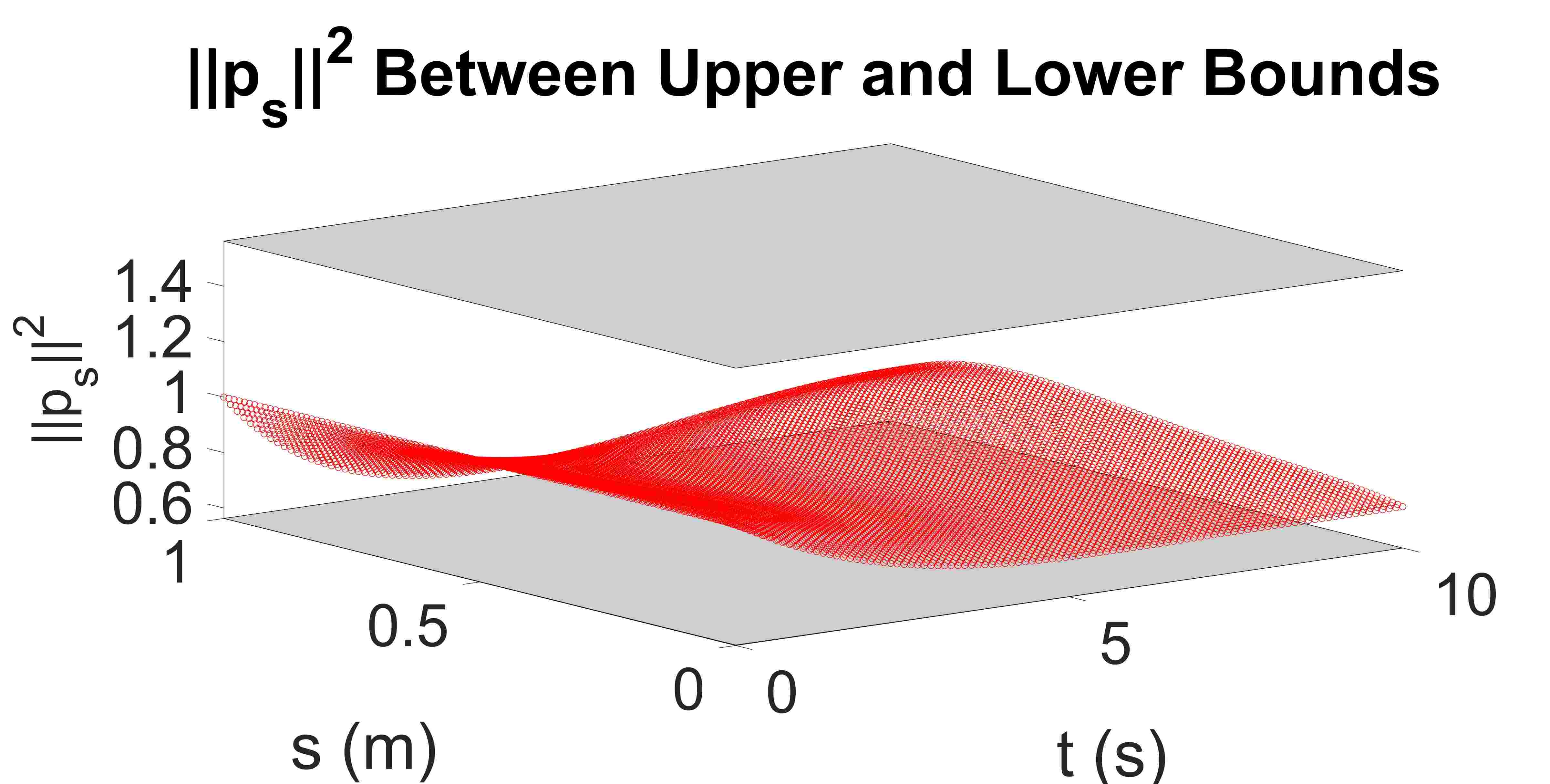}
        \caption{}
    \end{subfigure}
    \begin{subfigure}{.21\textwidth}
        \centering
        \includegraphics[width=\linewidth]{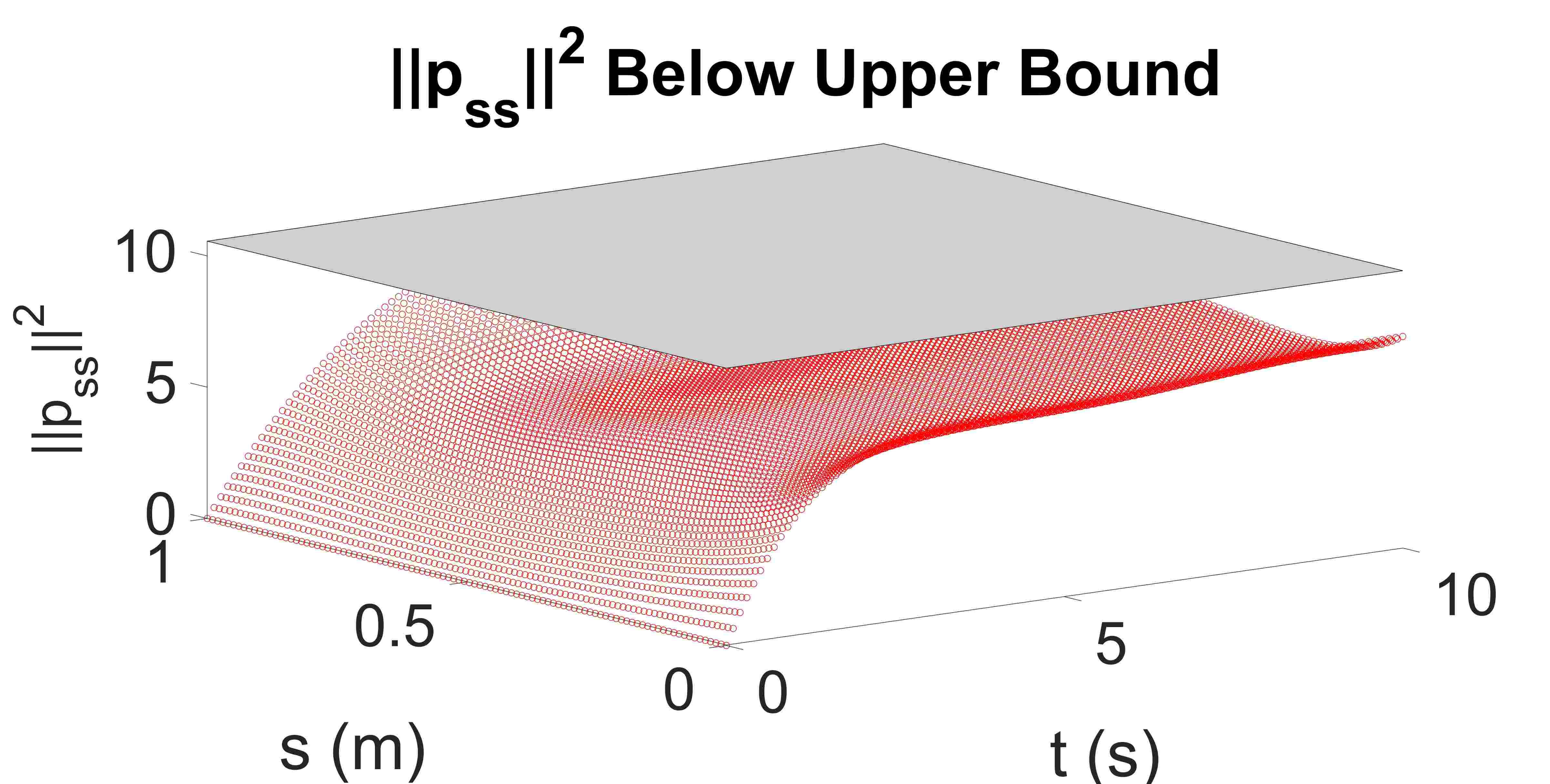}
        \caption{}
    \end{subfigure}
    \begin{subfigure}{.21\textwidth}
        \centering
        \includegraphics[width=\linewidth]{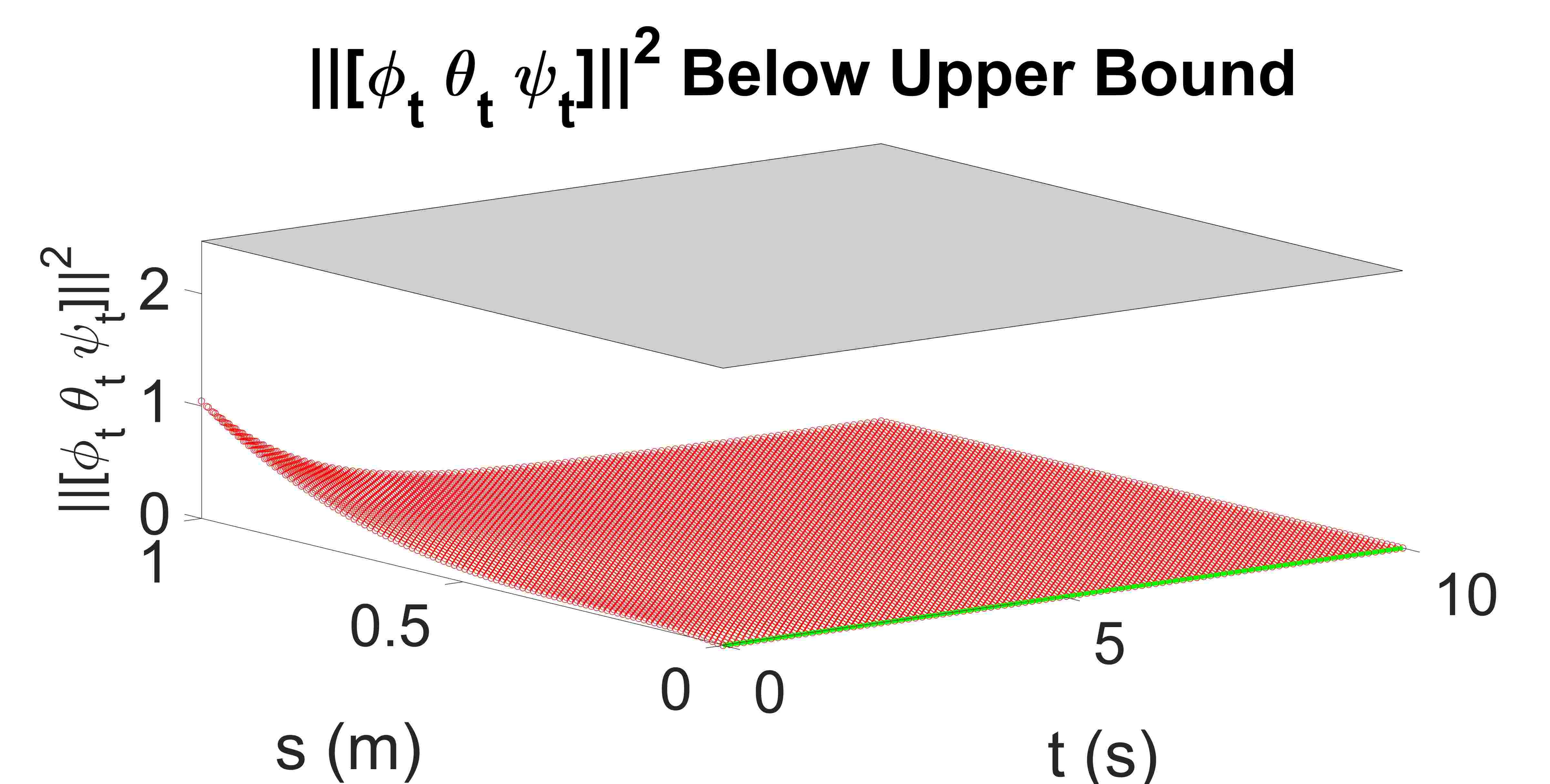}
        \caption{}
    \end{subfigure}
    \begin{subfigure}{.21\textwidth}
        \centering
        \includegraphics[width=\linewidth]{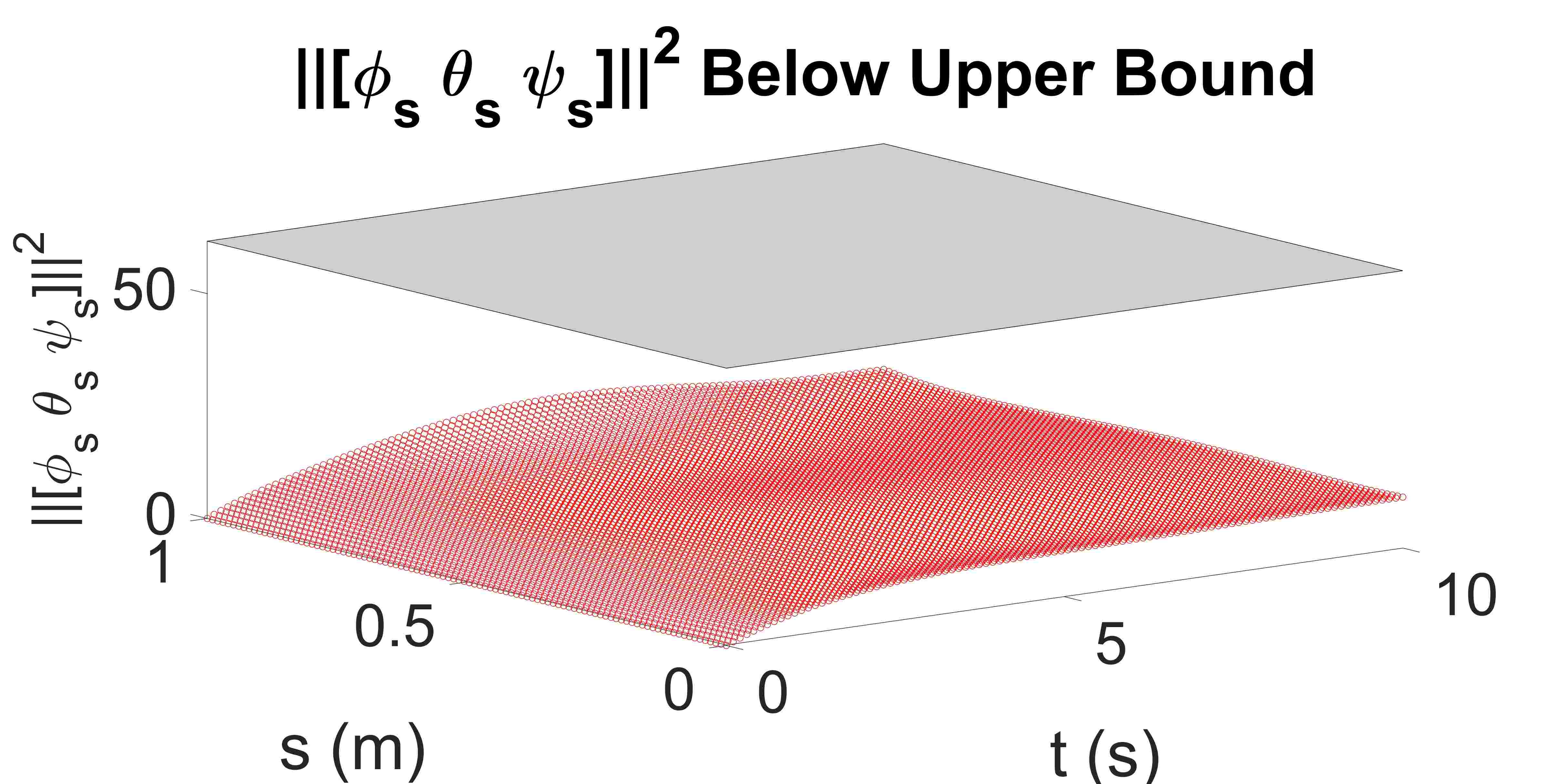}
        \caption{}
    \end{subfigure}
    \caption{Case 3: obstacle planning. Obstacles at [-0.115, 0.3, 0.65], [0.2, 0.2, 0.55] and [0.05, 0.25, 0.25] with 0.15, 0.13 and 0.20 $m$ radii respectively.}
    \label{fig:case3}
\end{figure}

Solutions are generated using fmincon in MATLAB on an Intel\textsuperscript{\textregistered} Core\textsuperscript{\texttrademark} i9-10885H CPU at 2.40GHz, 2400 Mhz with 8 Cores and 16 Logical Processors. In each case, a green line denotes the rod's initial configuration and a red line gives the terminal pose. A blue line is plotted along the trajectory's surface at every second to show the rod at that point in time. Blue circles denote the control points of the rod's position. A pink point and coordinate system plotted with thin lines show the desired final position and orientation of the rod's tip. A thicker coordinate system is plotted on the rod's tip. The surfaces demonstrating constraint satisfaction for all $s$ and $t$ is shown for each case. Figure \ref{fig:case1} shows a path for a case with no obstacles (Cases 1) starting with a straight initial configuration. Case 2 demonstrates path planning from a curved initial configuration maneuvering through two obstacles. Case 3 is shown in Figure \ref{fig:case3}, as an example of path planning to a desired pose between three spherical obstacles. Information about each case is summarized in Table \ref{table:case}. 

Increasing or decreasing the order of approximation within the problem can have significant impact on solution run times and results. Increasing order will approach the system's true optimal solution at the cost of longer run time. It is important to note that solutions are guaranteed safe even at lower orders through the properties of Bernstein polynomials \cite{kielas2022bernstein}. In cases where the desired position or orientation cannot be reached, the solution will be the minimum error of position and orientation based on the chosen optimization weights.

\begin{figure}
    \centering
    \begin{subfigure}{.4\textwidth}
        \centering
        \includegraphics[width=\linewidth]{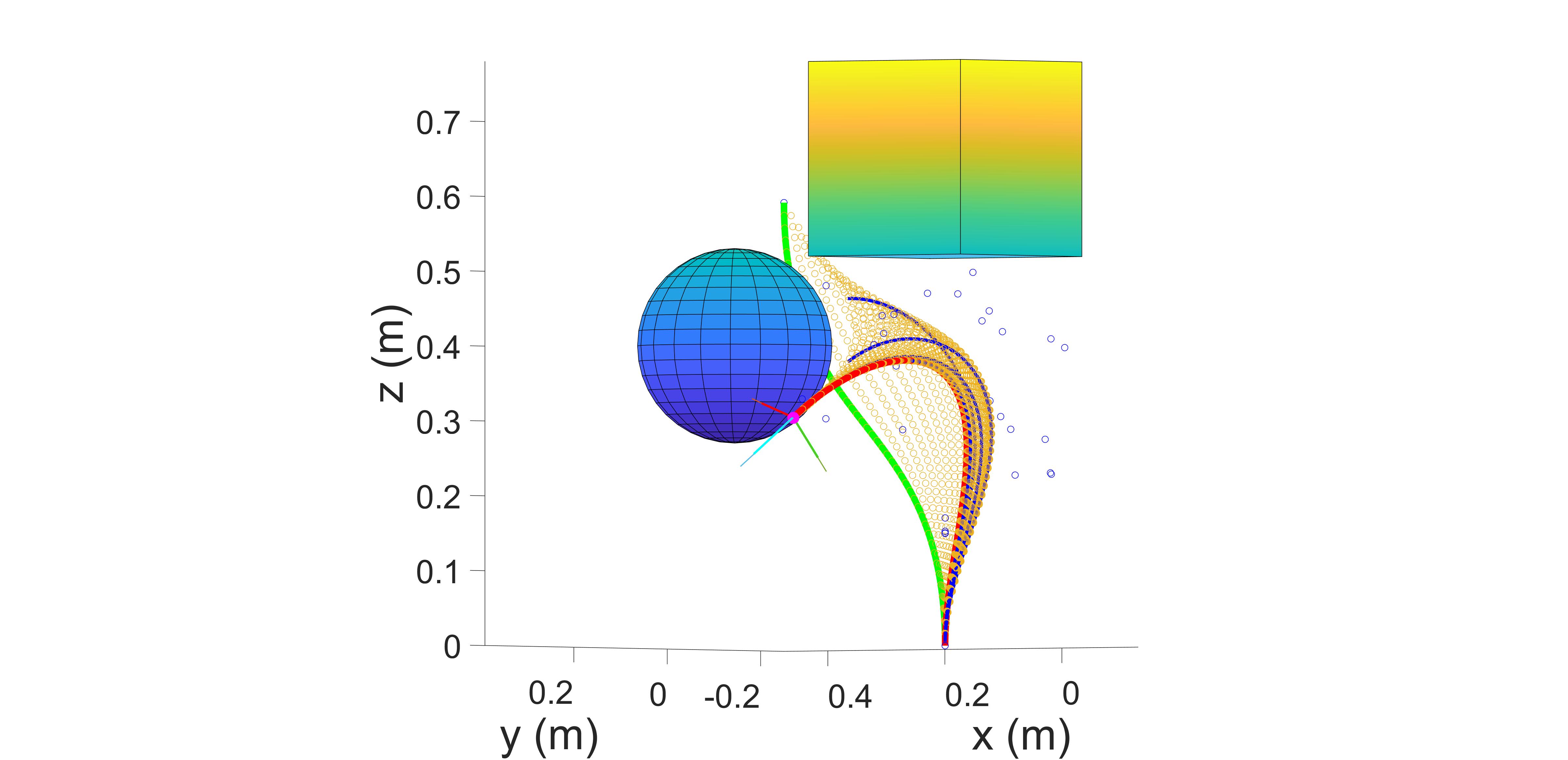}
        \caption{}
    \end{subfigure}
    \begin{subfigure}{.21\textwidth}
        \centering
        \includegraphics[width=\linewidth]{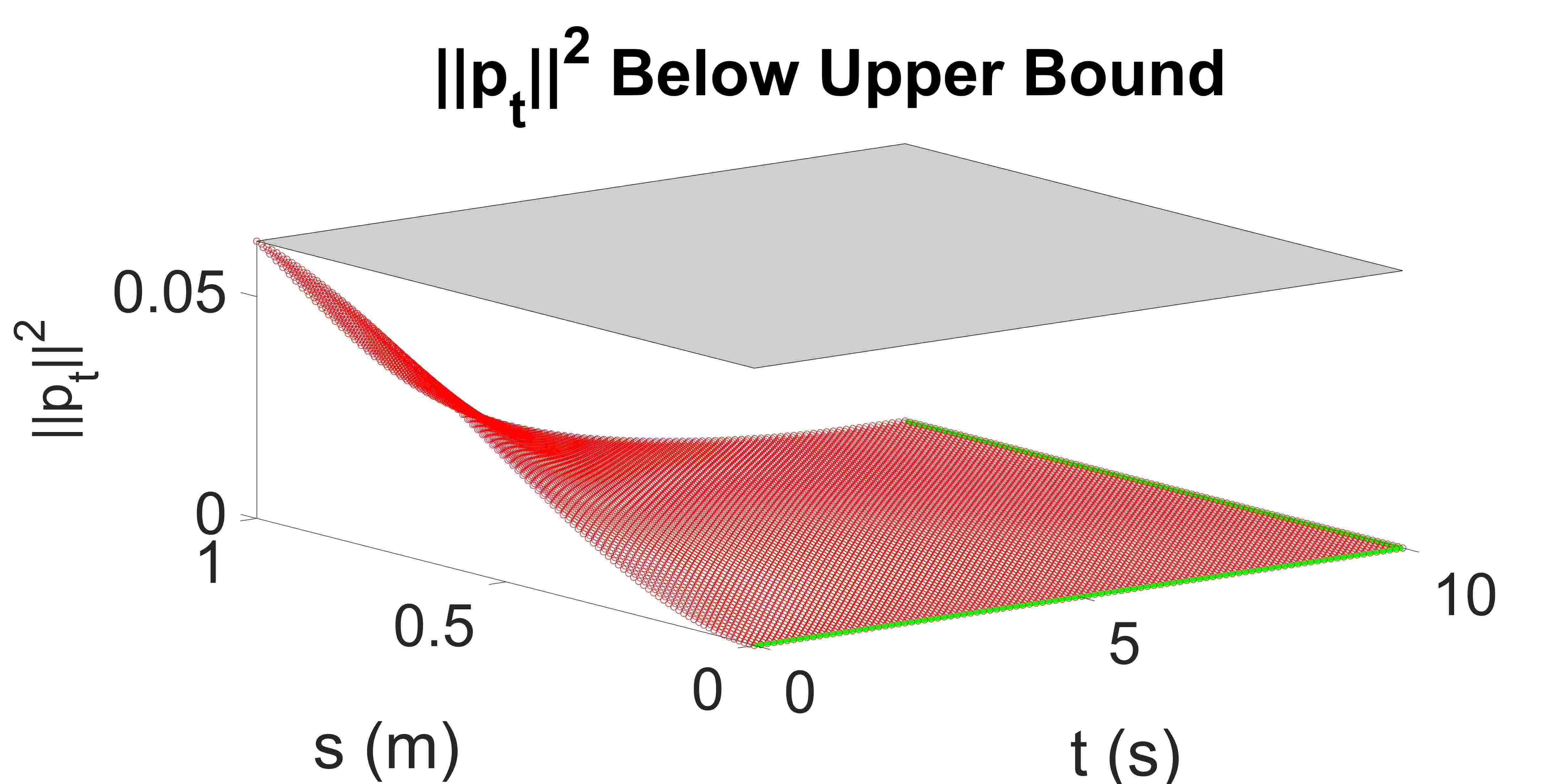}
        \caption{}
    \end{subfigure}
    \begin{subfigure}{.21\textwidth}
        \centering
        \includegraphics[width=\linewidth]{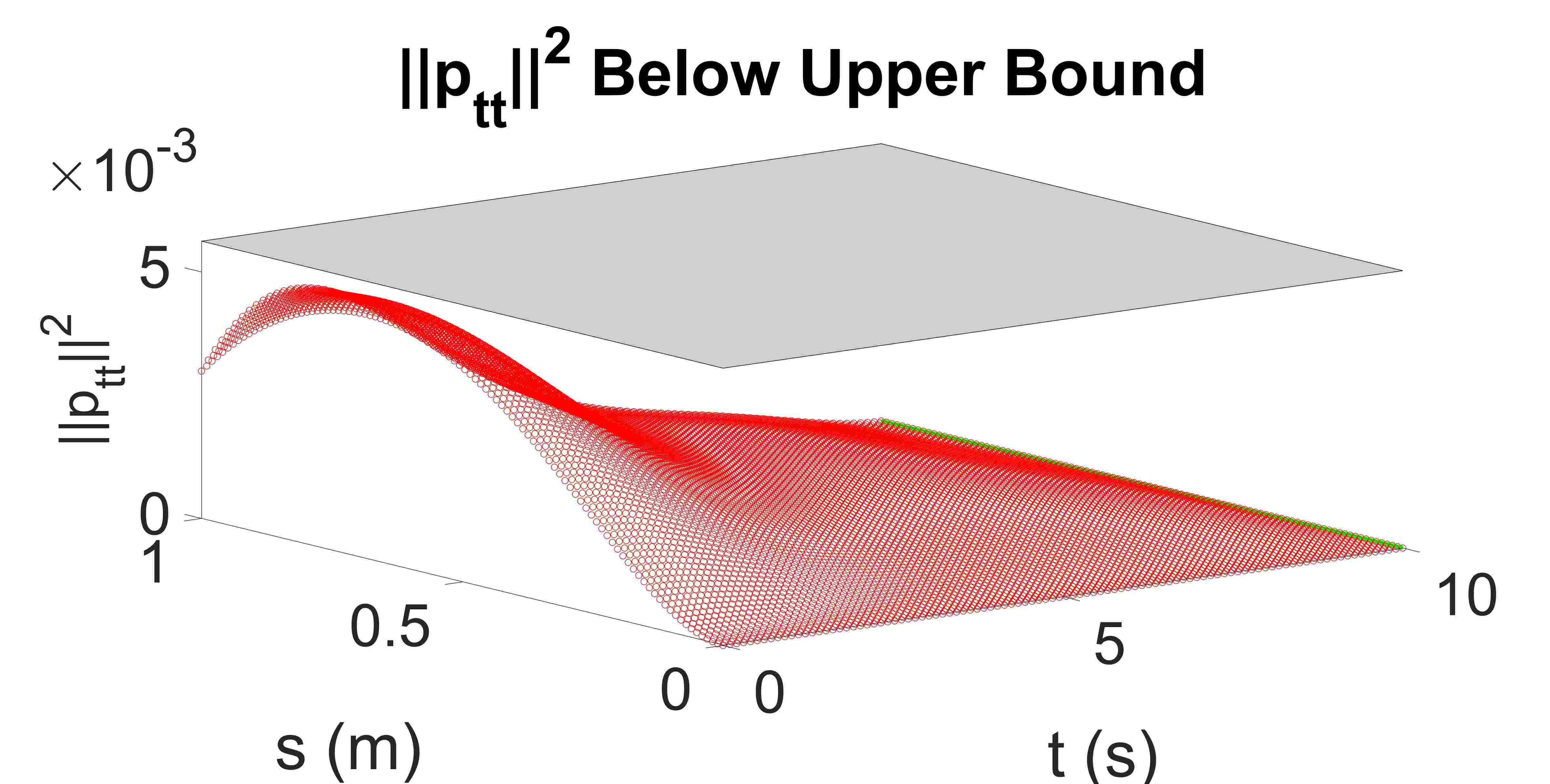}
        \caption{}
    \end{subfigure}
    \begin{subfigure}{.21\textwidth}
        \centering
        \includegraphics[width=\linewidth]{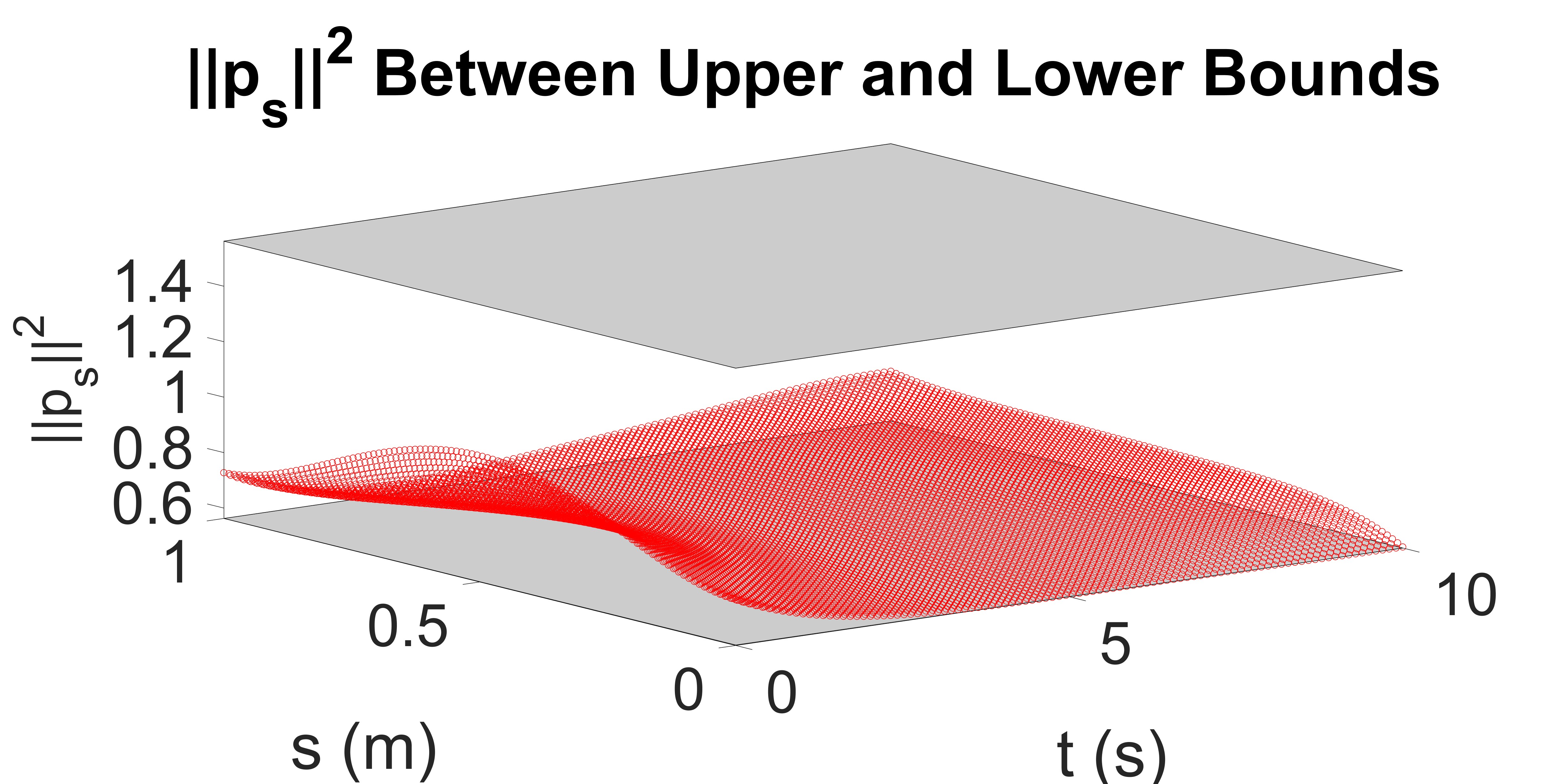}
        \caption{}
    \end{subfigure}
    \begin{subfigure}{.21\textwidth}
        \centering
        \includegraphics[width=\linewidth]{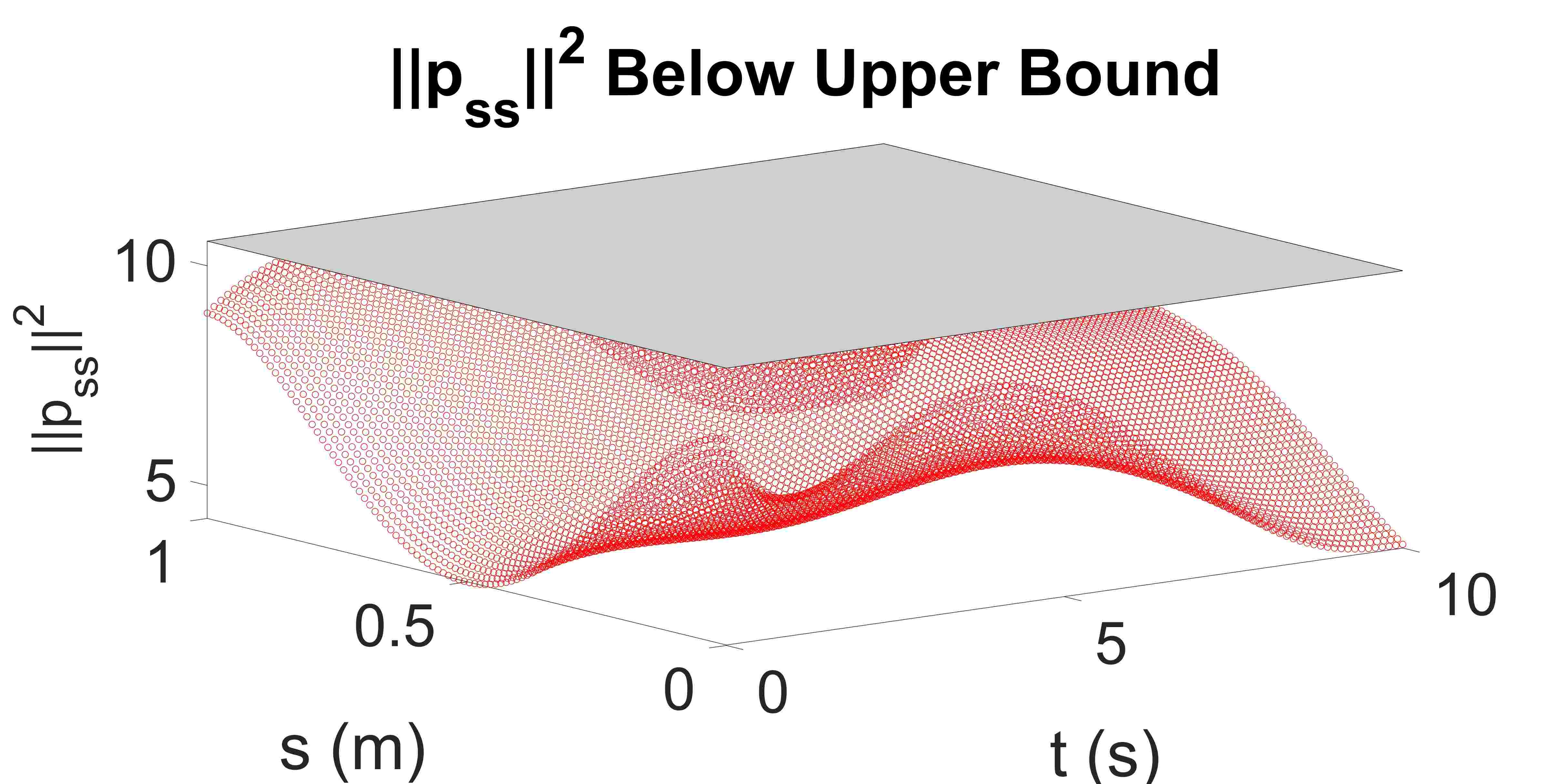}
        \caption{}
    \end{subfigure}
    \begin{subfigure}{.21\textwidth}
        \centering
        \includegraphics[width=\linewidth]{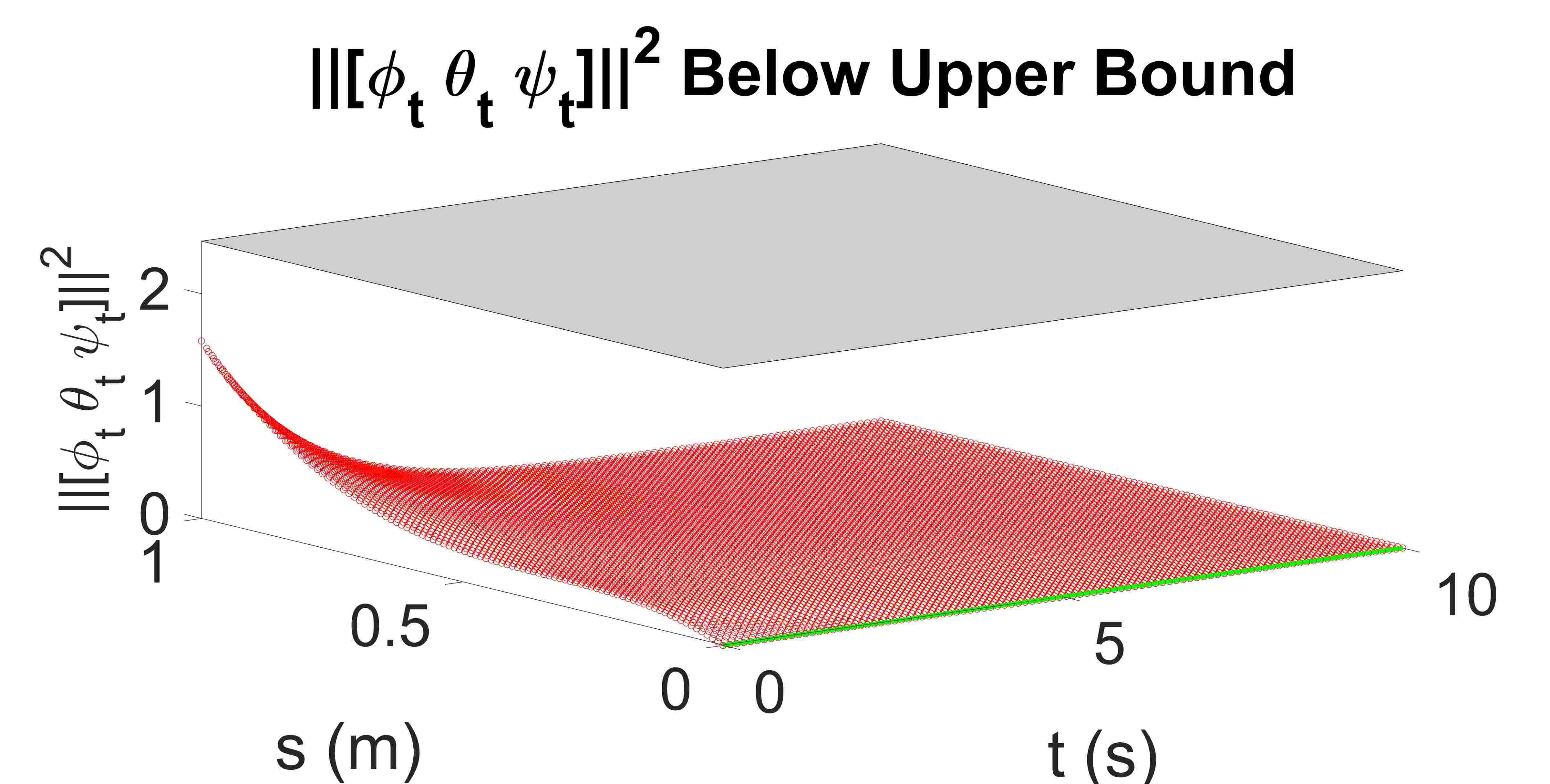}
        \caption{}
    \end{subfigure}
    \begin{subfigure}{.21\textwidth}
        \centering
        \includegraphics[width=\linewidth]{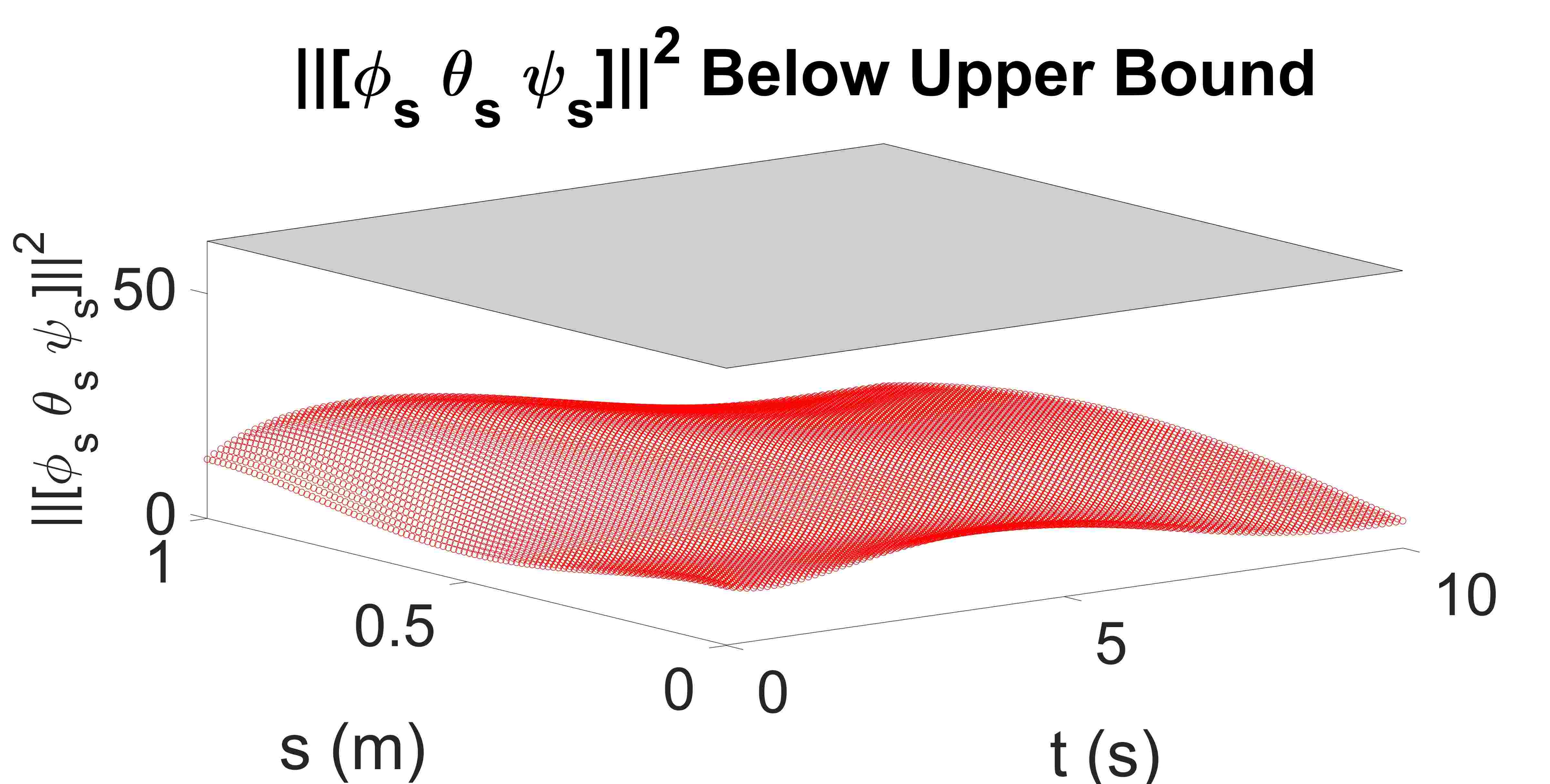}
        \caption{}
    \end{subfigure}
    \caption{Case 2: obstacle planning with curved initial pose. Obstacles at [0, 0, 0.65] and [0.2, 0.2, 0.4] with 0.13 $m$ edge length and radius respectively.}
    \label{fig:case2}
\end{figure}

\section{Conclusion}
\label{sec:conc}
In this paper, we presented a novel method to address the optimal motion planning problem for continuum rods by employing Bernstein surfaces. The main contribution is the approximation of continuous problems into their discrete counterparts, facilitating their solution using standard optimization solvers. This discretization leverages the unique properties of Bernstein surfaces, providing a framework that extends previous works which focused on ODEs approximated by Bernstein polynomials. Numerical validations were conducted through three distinct numerical scenarios. 

The presented methodology offers a promising direction for solving complex optimal control problems in the realm of continuum rod dynamics. Future works might delve deeper into the theoretical underpinnings, and further test the approach across a broader range of practical scenarios.

\bibliographystyle{IEEEtran}
\bibliography{refs}
\end{document}